\PassOptionsToPackage{table}{xcolor}
\documentclass[acmsmall]{acmart}

\usepackage{amsthm}
\theoremstyle{definition}

\usepackage{amsmath,amsfonts,mathtools} 
\usepackage{algorithm}
\usepackage{algorithmic}
\usepackage{graphicx}
\usepackage{textcomp}
\usepackage{color}
\usepackage{ctable}
\usepackage{array}
\usepackage{arydshln}

\newcolumntype{L}[1]{>{\raggedright\let\newline\\\arraybackslash\hspace{0pt}}m{#1}}
\newcolumntype{C}[1]{>{\centering\let\newline\\\arraybackslash\hspace{0pt}}m{#1}}
\newcolumntype{R}[1]{>{\raggedleft\let\newline\\\arraybackslash\hspace{0pt}}m{#1}}

\usepackage{booktabs} 
\usepackage{multirow}
\usepackage{colortbl}
\usepackage{subcaption}
\usepackage{bm}
\usepackage{verbatim}
\usepackage{comment}

\usepackage{enumitem}
\usepackage{tablefootnote}
\usepackage{url}

\newcommand{\ie}{\emph{i.e.}, }
\newcommand{\etal}{\emph{et al.} }
\newcommand{\eg}{\emph{e.g.}, }

\usepackage[switch]{lineno}
\usepackage{makecell}
\usepackage{array}

\AtBeginDocument{%
  \providecommand\BibTeX{{%
    \normalfont B\kern-0.5em{\scshape i\kern-0.25em b}\kern-0.8em\TeX}}}

\setcopyright{acmcopyright}
\copyrightyear{2022}
\acmDOI{10.1145/1122445.1122456}

\acmJournal{tmis}

\begin{document}

\title{Securing Visually-Aware Recommender Systems: An Adversarial Image Denoising and Detection Framework}

\author{Minglei~Yin}
\email{myin@albany.edu}
\affiliation{%
  \institution{University at Albany}
  \country{USA}
}

\author{Bin~Liu}
\email{bin.liu1@mail.wvu.edu}
\orcid{0000-0002-7028-5334}
\affiliation{%
  \institution{West Virginia University}
  \country{USA}
}

\author{Neil Zhenqiang~Gong}
\email{neil.gong@duke.edu}
\affiliation{%
  \institution{Duke University}
  \country{USA}
}

\author{Xin~Li}
\email{xli48@albany.edu}
\affiliation{%
  \institution{University at Albany}
  \country{USA}
}

\renewcommand{\shortauthors}{Yin, et al.}

\begin{abstract}
With rich visual data, such as images, becoming readily associated with items, visually-aware recommendation systems (VARS) have been widely used in different applications.  Recent studies have shown that VARS are vulnerable to item-image adversarial attacks, which add human-imperceptible perturbations to the clean images associated with those items. Attacks on VARS pose new security challenges to a wide range of applications, such as e-commerce and social media, where VARS are widely used. How to secure VARS from such adversarial attacks becomes a critical problem. Currently, there is still a lack of systematic studies on how to design defense strategies against visual attacks on VARS. 
In this paper, we attempt to fill this gap by proposing an \emph{adversarial image denoising and detection} framework to secure VARS. Our proposed method can simultaneously (1) secure VARS from adversarial attacks characterized by \emph{local} perturbations by image denoising based on \emph{global} vision transformers; and (2) accurately detect adversarial examples using a novel contrastive learning approach.  Meanwhile, our framework is designed to be used as both a filter and a detector so that they can be \emph{jointly} trained to improve the flexibility of our defense strategy to a variety of attacks and VARS models. 
Our approach is uniquely tailored for VARS, addressing the distinct challenges in scenarios where adversarial attacks can differ across industries, for instance, causing misclassification in e-commerce or misrepresentation in real estate.
We have conducted extensive experimental studies with two popular attack methods (FGSM and PGD). Our experimental results on two real-world datasets show that our defense strategy against visual attacks is effective and outperforms existing methods on different attacks.  Moreover, our method demonstrates high accuracy in detecting adversarial examples, complementing its robustness across various types of adversarial attacks.
\end{abstract}



\maketitle
\pagestyle{plain}

\def \Pr {{\mathrm{Pr}}}
\def \tr {\mathrm{tr}}

\def \xx {{\bf x}}
\def \XX {{\bf X}}

\def \YY {{\bf Y}}

\def \ZZ {{\bf Z}}

\def \ww {{\bf w}}
\def \WW {{\bf W}}

\def \pp {{\bf p}}
\def \PP {{\bf P}}

\def \qq {{\bf q}}
\def \QQ {{\bf Q}}

\def \uu {{\bf u}}
\def \UU {{\bf U}}

\def \vv {{\bf v}}
\def \VV {{\bf V}}

\def \bb {{\bf b}}
\def \BB {{\bf B}}

\def \GG {\bf G}
\def \RR {{\bf R}}

\def \II {{\bf I}}

\section{Introduction}

In the era of data explosion, people often face overwhelming information overload problems. Recommender systems play an important role in helping users find the information they are interested in more easily \cite{adomavicius2005toward}. As of today, recommender systems have become essential components in a wide range of Internet services---from E-commerce to social networks---to help users deal with information overload, engage users, and improve user experience. 
In a typical recommender system setting, we are given a set of users, a set of items, and a record of the users' historical interactions (e.g., ratings, likes, or clicks) with the items, and our goal is to model user preferences from the user-item interactions and then recommend each user a list of new items that the users have not experienced yet.
Meanwhile, as rich visual data, such as images, are becoming more widely associated with items, visually aware recommender systems (VARS)  have been widely used in several application domains such as e-Commerce \cite{he2016vbpr,kang2017visually}, image sharing and social networks \cite{pal2020pinnersage,zhai2017visual, Wang:POI:www17}, fashion \cite{hu2015collaborative}, food \cite{elsweiler2017exploiting} and real estate and tourism \cite{sertkan2020pictoure}. As the cliché says, ``a picture is worth a thousand words'' -- the visual appearance of a product image (e.g., the picture of an outfit or an apartment) could affect the final decision of an online consumer \cite{he2016ups,he2016vbpr}. 
However, the rise of adversarial image attacks poses a critical threat to these systems. For example, in e-Commerce, adversaries may subtly alter product images to manipulate rankings, potentially causing misdirected purchasing decisions and undermining brand integrity. Similarly, in real estate, compromised property images might distort market perceptions, leading to inaccurate valuations and misguided investments.

Recent studies have shown that visually aware recommender systems (VARS) are vulnerable to item image adversarial attacks, which add human-imperceptible perturbations to clean images associated with these items \cite{tkde20amr,di2020taamr,liu2020adversarial,Cohen2021AttackVisually}. Tang et al. \cite{tkde20amr} found that a small but intentional perturbation in the input image will severely decrease the accuracy of the recommendation, implying the vulnerability of VARS to untargeted attacks. 
Noia et al. \cite{di2020taamr} studied targeted attacks to VARS by perturbing images of low recommended category of products to be misclassified as a more recommended category.
A black-box attack model for VARS was studied  in \cite{Cohen2021AttackVisually} and it shows that the visual attack model can effectively influence the preference scores and classifications of items without knowing the parameters of the model to promote the push of certain items.  Liu et al. \cite{liu2020adversarial} studied adversarial item promotion attacks at VARS in the top N item generation stage in the cold start setting.  
Although there exist many studies \cite{lam2004shilling,xing2013take,li2016data,fang2018poisoning,chen2020data,song2020poisonrec,liu2020certifiable,huang2021data,zhang2021reverse,wu2021triple,xu2023vulnerability,nguyen2023poisoning} on attacks on general recommendation systems that manipulate user-item interaction data in different ways, attacks against VARS are different in that they only manipulate images associated with items. The security aspect of VARS systems is much less explored.

Attacks on VARS pose new security challenges to a wide range of applications such as e-Commerce and social media where VARS are widely used. 
However, to our knowledge, there has been a limited amount of research on defending VARS from adversarial attacks. Tang \etal \cite{tkde20amr} proposed to apply adversarial training to improve the robustness of VARS. 
Anelli \etal \cite{ADDMMSIGIR21} conducted a study on the effectiveness of adversarial training methods to improve the robustness of VARS to different adversarial image manipulations including the Fast Gradient Sign Method (FGSM) \cite{goodfellow2014explaining}, Projected Gradient Descent (PGD) \cite{kurakin2016adversarial},  and Carlini \& Wagner (CW) \cite{carlini2017towards} attack. Despite the fact that there are various studies on attacks and defenses on vision learning systems \cite{goldblum2022dataset,Adversarial2019TNNLS}, there is still no systematic study on the defense of VARS against increasingly more powerful attacks.  Generally speaking,   \emph{robust model construction} and  \emph{attack detection}  are two popular strategies for defending against attacks against recommender systems \cite{deldjoo2021survey}. 
Robust model construction approaches, such as robust statistics-based methods \cite{mehta2007robust} and more recently adversarial training-based methods \cite{tkde20amr,ADDMMSIGIR21,he2018adversarial,yuan2019adversarial,wu2021fight}, aim to design recommender systems proactively that
are more secure against attacks. Attack detection approaches aim to detect malicious user profiles \cite{burke2006classification, wu2012hysad,mehta2009unsupervised,lee2012shilling} and recover from attacks.

In this paper, we address this gap by presenting a novel \emph{adversarial image denoising and detection} framework to protect visually aware recommender systems from adversarial attacks that manipulate images associated with items. The proposed framework is designed to simultaneously (1) mitigate the impacts of adversarial attacks on recommendation performance and (2) detect adversarial attacks. 
Specifically, the adversarial image denoising component reconstructs clean item images by denoising the adversarial perturbations in attacked images. To this end, we develop an image denoising network composed of several residual blocks and a global vision transformer \cite{paul2021vision}. Since adversarial attacks such as FGSM and PGD are local perturbations, the proposed transformer-based global filtering strategy can effectively make the denoised images as close to clean images as possible, alleviating the adversarial impact on the performance of VARS models. 
Moreover, we also design a novel contrastive learning-based component to accurately detect adversarial examples. A novel strategy is proposed to construct positive and negative pairs for contrastive learning. By pushing the denoised images toward clean images and away from adversarial ones, our detection network can detect adversarial examples with high accuracy. 
Furthermore, our framework is designed to be a flexible filter and detector plug-in, which can defend against the adversarial attack without modifying the original recommender system. We build our defense framework on Visual Bayesian Personalized Ranking (VBPR)~\cite{he2016vbpr}, the most popular VARS model, as our baseline model. 

We evaluate our defense framework with two popular attack methods, \ie FGSM and PGD. Our experimental results on two real-world datasets (Amazon Men and Amazon Fashion) show that our framework can effectively defend against visual attacks and outperforms existing methods on different attacks. Furthermore, our method can detect adversarial examples with high accuracy. 

A summary of key contributions is listed below.
\begin{itemize}
    \item We provide a systematic study on securing visually-aware recommender systems (AVRS) by unifying two defense strategies---\emph{robust model construction} and \emph{attack detection}---from adversarial attacks that manipulate images associated with items.   Consistent with the knowledge contribution framework (KCF) \cite{samtani2023deep}, our study contributes to the emerging literature on AI security by introducing an framework to defense against adversarial attacks to AVRS. 
    \item We design a novel \emph{adversarial image denoising and detection} framework to simultaneously mitigate the impacts of adversarial attacks and detect the attacks.  We demonstrate that end-to-end training of denoising and detection networks can significantly improve the robustness of VARS to a variety of adversarial attacks.
  \item  Extensive experiments on two real-world datasets demonstrate that the proposed framework can effectively defend the recommender system model from attacked images with varying strengths. Moreover, our proposed framework can detect adversarial examples with high accuracy.
\end{itemize}

\section{Related Work}
We organize the related work into two categories: security of general recommender systems, and attacks and defenses in visually-aware recommender systems. 

\subsection{Security of General  Recommender Systems}

In the era of data explosion, recommender systems play a crucial role in reducing information overload by helping users find relevant content efficiently \cite{adomavicius2005toward}. In a \emph{general recommender system} setting, we are given a set of users, a set of items, and a record of the users' historical interactions (e.g., ratings, likes, or clicks) with the items, and our goal is to model user preferences from the user-item interactions and then recommend to each user a list of new items that the users have not experienced yet. Many algorithms have been developed for this purpose, including neighborhood-based~\cite{sarwar2001item}, matrix-factorization-based~\cite{koren2009matrix, mnih2007probabilistic}, graph-based~\cite{fouss2007random}, and deep-learning-based approaches~\cite{he2017neural,zhang2019deep,wu2022graph}.

While many methods have been proposed to improve recommendation performance, the security aspect of recommender systems is much less explored but has received increasingly more attention in recent years \cite{deldjoo2021survey}. 
Due to the nature of openness, where user-item interaction data is used to train a recommendation system,  a body of studies has shown that recommender systems are
vulnerable to various adversarial attacks \cite{lam2004shilling,xing2013take,li2016data,fang2018poisoning,chen2020data,song2020poisonrec,liu2020certifiable,huang2021data,zhang2021reverse,wu2021triple,xu2023vulnerability,nguyen2023poisoning}, such as data poisoning or profile pollution, where attackers inject manipulated data to influence recommendations. Data poisoning involves injecting fake user profiles to promote or demote specific items \cite{lam2004shilling,li2016data,fang2018poisoning,chen2020data,song2020poisonrec,fang2020influence,huang2021data,zhang2021reverse,wu2021triple,xu2023vulnerability,nguyen2023poisoning}, while profile pollution aims to skew user profiles, potentially leading to incorrect recommendations \cite{xing2013take,liu2020certifiable}.

\emph{Robust model construction} and \emph{attack detection} are two major strategies to defend against attacks on recommendation systems \cite{deldjoo2021survey}. The first strategy is to proactively design robust recommender systems so that they are more secure against attacks.  
Alone this line, adversarial training \cite{ADDMMSIGIR21,he2018adversarial,yuan2019adversarial,wu2021fight} has been applied to improve the robustness of recommender systems. The basic idea of adversarial training \cite{goodfellow2014explaining} is to train a recommender system model on a training dataset that is augmented with adversarial examples so that the adversarially trained recommendation model is resistant to adversarial attacks. 
For example, He \etal \cite{he2018adversarial} proposed an adversarial personalized ranking framework that applied adversarial training to the widely used Bayesian Personalized Ranking (BPR) model \cite{BPR:2009} by introducing adversarial perturbations in the embedding vectors of users and items. Yuan \etal  \cite{yuan2019adversarial}  studied adversarial training on collaborative denoising auto-encoder recommendation model. Unlike the adversarial training framework in \cite{he2018adversarial,yuan2019adversarial} that add perturbations to the model parameters (\eg embedding vectors of users and items), Wu \etal \cite{wu2021fight} proposed an adversarial poisoning training method to counteract data poisoning attacks to  recommender systems.
The second strategy, the attack detection-based method, aims to detect malicious user profiles and then remove compromised user profiles in the data processing stage. 
Attack detection-based method assumes that malicious users and genuine users have different user-item interaction patterns. Different attack detection methods have been proposed, including classification \cite{burke2006classification,mobasher2007toward} by extracting attributes derived from user profiles and items, semi-supervised learning \cite{wu2012hysad}, and unsupervised learning such as clustering in the user-item rating matrix \cite{mehta2009unsupervised,lee2012shilling} and graph-based methods \cite{yang2016estimating} on user-item graph.  Zhang \etal \cite{zhang2020gcn} proposed a  method of unifying the robust recommendation task and fraudster detection task by combining a graph convolutional network (GCN) model to predict user-item ratings and a neural random forest model to predict the mean square of all ratings per user. They assumed that if a user's rating is
largely deviated from the predicted ratings, this user is most likely to be a fraudster. 

\subsection{Attacks and Defenses in Visually-Aware Recommender Systems} 
With rich visual data, such as images, becoming readily associated with items, visually aware recommendation systems (VARS) have been widely used to improve users' decision-making process and support online sales \cite{he2016vbpr,kang2017visually,he2016ups}. Typically, VARS models ~\cite{he2016vbpr,kang2017visually}  first extract image features using deep neural networks and then combine the extracted features with existing recommendation models \cite{deldjoo2021study}. 
Visual Bayesian Personalized Ranking (VBPR)~\cite{he2016vbpr} is one of the most widely used models in VARS. Specifically, VBPR uses neural networks such as pre-trained CNN such as $ResNet50$ \cite{he2015deep}  to extract image features from images associated with items and then fuses the extracted features into the widely used Bayesian personalized ranking (BPR) model \cite{BPR:2009}. 
More recently, visual-aware deep Bayesian personalized ranking (DVBPR)~\cite{kang2017visually} was developed to
simultaneously extract task-guided visual features and learn user latent factors, leading to improved recommendation performance by directly learning ``fashion-aware'' image representations through joint training of image representation and recommender systems. 

Recent studies have shown that VARS are vulnerable to item image adversarial attacks, which only add human-imperceptible perturbations to clean images associated with those items \cite{tkde20amr,di2020taamr,Cohen2021AttackVisually,liu2020adversarial}. 
Tang et al. \cite{tkde20amr} studied the vulnerability of VARS in an untargeted attack environment and found that a small but intentional perturbation in the input image will severely decrease the precision of the recommendation. 
Noia \etal \cite{di2020taamr} proposed a targeted attack to VARS, and formulated the attack goal as to spoof the recommender system to misclassify the images of a category of low recommended products towards the class of more recommended products. 
Since the item catalogs of VARS are usually large, recommender systems often count on the item providers to provide images as supplementary information. This reliance on external sources has inspired the design of a black-box attack on VARS in \cite{Cohen2021AttackVisually}. An attacker was shown to unfairly promote targeted items by modifying the item scores and pushing their rankings. By systematically creating human-imperceptible perturbations of the images of the pushed item, the attackers manage to incrementally increase the item score. 
Liu et al. \cite{liu2020adversarial} studied adversarial item promotion attacks in VARS in the top-N item generation stage under the cold-start recommendation setting.
Note that attacks on VARS are different from existing work on attacks against general recommendation systems \cite{lam2004shilling,li2016data,fang2018poisoning,chen2020data,song2020poisonrec,fang2020influence,huang2021data,wu2021triple,xu2023vulnerability,nguyen2023poisoning}. While studies on attacks against general recommender systems focus on manipulating user-item records to achieve attack goals, attacks against VARS only manipulate images associated with items.  

There is a limited amount of research on the defenses against item image adversarial attacks to VARS. After demonstrating the vulnerability of VARS in an untargeted attack environment, Tang et al. \cite{tkde20amr} proposed to apply adversarial training to improve the robustness of VARS. Specifically,  adversarial perturbations were applied to the item image's deep feature vector and the adversarial training was formulated as an adversarial regularizer added to the BPR loss. 
Anelli et al. \cite{ADDMMSIGIR21} conducted a study on the effectiveness of adversarial training methods to improve the robustness of VARS to different adversarial image manipulations, including the Fast Gradient Sign Method (FGSM) \cite{goodfellow2014explaining}, Projected Gradient Descent (PGD) \cite{kurakin2016adversarial},  and Carlini \& Wagner (CW) \cite{carlini2017towards} attack.  They assumed that adversaries were aware of recommendation lists, and then formulated the adversarial attack as to misclassify the images
of a category of low-recommended products towards the class of more recommended products.
Instead of focusing on the item-level raking performances of recommender systems, models were evaluated in terms of the fraction of compromised items in the top-$N$ recommendations. 
Our work is different from previous research \cite{tkde20amr,ADDMMSIGIR21} on defenses against item image adversarial attacks to VARS in several different ways. 
First,  both our work and \cite{tkde20amr} consider untargeted attacks to VARS with the goal of dysfunctioning the recommender system; however, the ways to generate adversarial samples are different. In \cite{tkde20amr} perturbations are added to model parameters (\ie item image's deep feature vector), but the  attack in our study is to add human-imperceptible perturbations to clean images associated with items. 

Second, from the defense perspective,  \cite{tkde20amr,ADDMMSIGIR21}  only applies adversarial training--a robust model construction strategy--to boost the robustness of the recommendation. In contrast to the existing studies, our proposed \emph{adversarial image denoising and detection} method combines two defense strategies--robust model construction and attack detection--in a framework. Specifically,  the denoising network enhance the robustness of the recommendation through generating clean images, and the detection network detects adversarial input by  separating adversarial inputs from clean inputs. As demonstrated in  the
empirical study, two networks play different roles in the defense against adversarial attack signals, and they are mutually beneficial to each other.
Moreover, beyond boosting the robustness of VARS, our framework is able to detect compromised item images, which would have important practical implications.  
We  note that orthogonal to VARS,  attacks and their defenses have also been studied in various image classification algorithms in the computer vision community \cite{goldblum2022dataset, Adversarial2019TNNLS, tian2018detecting,  liang2018detecting, wang2023addition}. 
Traditional adversarial defenses in computer vision—often focusing on classification tasks—do not readily translate to the regression-based challenges in VARS.

\section{Proposed Method}

In this section, we first introduce item image adversarial attacks to visually aware recommendation systems in this study. We then introduce our proposed framework, which takes into account both \emph{robust model construction} and \emph{attack detection} into account, to defend against such attacks and elaborate details about our proposed defense method. 

\subsection{Preliminaries and Problem Formulation}
\label{sec:3.1}

Our goal is to secure visually-aware recommendation systems (VARS) from adversarial attacks. 
Specifically, we consider a typical VARS setting where we have a set of $M$ users $\mathcal U =\{1, 2, \dots, M\}$ and a set of $N$ items $\mathcal I=\{1, 2, \dots, N\}$, and we are given a record of user-item interactions $\mathcal{D}=\{\langle u, i, r_{ui} \rangle\}$, where $r_{ui}$ denotes the preference score of user $u$ for item $i$. We assume that an image $x_i$ is associated with each item $i$, and the images are denoted as $\mathcal{X}=\{x_i\}_{i\in \mathcal{I}}$. In this paper, without loss of generality, we focus on the widely used VARS model, Visual Bayesian Personalized Ranking (VBPR)~\cite{he2016vbpr}, as our baseline model, although the methodology could be generalized to other VARS. To avoid overfitting, following \cite{tkde20amr} we define the user preference score of user $u$ for the item $i$ as ${r}_{ui}=\gamma_{u}^{T}\left(\gamma_{i}+ E f_{i}\right)$,
where $\gamma_{u}$ and $\gamma_{i}$ are latent factors of the user and the item, respectively, $f_{i}$ is the vector of visual features extracted by a pre-trained CNN such as $ResNet50$ \cite{he2015deep} from the image $x_i$ associated with the item $i$, and $E$ is a transformation matrix that maps the visual feature $f_{i}$ into the latent factor space of the item. Then the task of VARS is to build a model $r_{ui}=F(u, i, x_i|\Theta)$, where $u\in \mathcal U$, $i \in \mathcal I$, and $\Theta$ indicate the model parameters, to infer the preference scores for the items that the users have not yet experienced.  Recommendations are made based on the inferred preference scores. We use Bayesian Personalized Ranking (BPR) optimization framework, which is a pairwise ranking loss, to train the VBPR model. Specifically, we first construct a training dataset $\mathcal{D}_{s}$ as follows: 
\begin{equation} \label{eq:dataset}
\mathcal{D}_{s}=\left\{(u, i, j, x_i, x_j) \mid u \in \mathcal{U} \wedge i \in \mathcal{I}_{u}^{+} \wedge j \in \mathcal{I} / \mathcal{I}_{u}^{+}\right\},
\end{equation}
where $\mathcal{I}_{u}^{+}$ represents the set of interacted items of user $u$. The triplet $(u,i,j)$ indicates that the user $u$ prefers the positive item $i$ over the negative item $j$. We uniformly sample the negative item $j$  from the set $\mathcal{I} / \mathcal{I}_{u}^{+}$, representing items that user $u$ has not previously interacted with. Here, $x_i$ and $ x_j$ represent the images of items $i$ and $j$, respectively. BPR loss is defined as follows:
\begin{equation}\label{Equ:bpr}
        \mathcal{L}_{BPR} = \underset{\Theta}{\operatorname{argmin}}  \left\{ - \sum\limits_{(u,i,j) \in \mathcal{D}_s} \ln{\sigma (r_{ui}-r_{uj})} + \lambda \| \Theta \|^2 \right\},
\end{equation}
where $\sigma(\cdot)$ is the Sigmoid function and $\lambda$ is a regularization hyperparameter. \vspace{5pt}

{\bf \noindent Untargeted Attacks to VARS}. The goal of untargeted attacks to VARS is to  disrupt  the recommender system. Specifically, the objective of adversarial attacks to VARS is  to alter the ranking of item recommendations generated by the VARS, by adding human-imperceptible perturbations $\delta_i$ to clean images $x_i$ associated with each item $i \in \mathcal{I}$, namely, $x_{i}^{*} = x_{i} + \delta_{i}$. 
We consider a white-box attack setting in which we assume that an attacker has access to the user-item interaction data $\mathcal D$ and the images $x_{i}$ associated with the item $i$, the recommender system model (the VBPR model in this work) and the neural network architecture to derive image features.
When  a user $u$ prefers item $i$ over item $j$, then adversarial attacks aim to alter preference scores $F(u, i, x_i|\Theta)$ and $F(u, j, x_j|\Theta)$ so that $F(u, i, x_i|\Theta)   < F(u, j, x_j|\Theta)$.
Meanwhile, to make perturbations imperceptible, the attacked image $x_{i}^{*}$ ($x_{j}^{*}$) should be similar to the clean image $x_{i}$ ($x_{j}$). 
The generation of adversarial samples can be modeled as a constrained minimization problem:
\begin{equation}\label{Equ:attack}
\begin{split}
  \text{minimize } \;& \|x_i^*-x_i\|_2^2 + \|x_j^*-x_j\|_2^2 \\
 \;&  + F(u, i, x_i|\Theta)   - F(u, j, x_j|\Theta) \\
  \text{such that } \;&  x_i^*,x_j^* \in [0,1]^n,
\end{split}
\end{equation}
where $i$ and $j$ indicate positive and negative samples in the VBPR model, respectively. 
The first part $\|x_i^*-x_i\|_2^2 + \|x_j^*-x_j\|_2^2$ aims to keep the attacked images $x_i^*$ and $x_j^*$ to stay close to the original images $x_i$ and $x_j$ in pixel space, respectively, 
the second part $F(u, i, x_i|\Theta)   - F(u, j, x_j|\Theta) $ disrupts the output of VARS by flipping the preference scores, 
and $x_i^*,x_j^* \in [0,1]^n$ are image value range constraints to keep the perturbed images within the original image value range.

\begin{figure}[!t]
\centering
\includegraphics[width=0.8\textwidth]{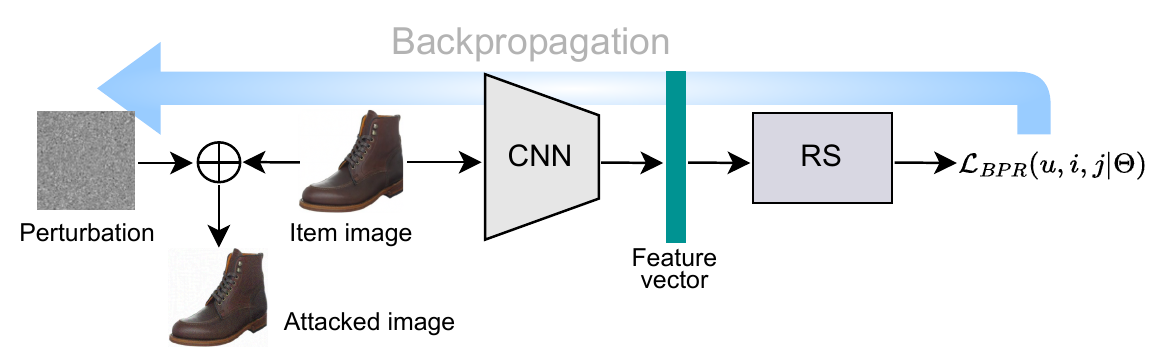}
\caption{Illustration of adversary example generation in attacks to visually-aware recommender systems. Adversary examples are generated based on the VBPR model ~\cite{he2016vbpr}, which includes a pre-trained CNN (\eg ResNet50) for image feature extraction and a latent factor recommendation model for user preference prediction. The whole pipeline is differentiable. Therefore, the attack signal can be computed by the backpropagation of the loss value.\label{fig:attack}}
\end{figure}

Given the adversarial attack goal as shown in Equation (\ref{Equ:attack}), different attack methods, such as FGSM \cite{goodfellow2014explaining} and PGD \cite{kurakin2016adversarial}, can be used to generate adversarial samples,  and the perturbed
images are denoted as ${\mathcal{X}}^*=\{x_i^*\}_{i\in \mathcal{I}}$. Specifically, the preference score flipping part $F(u, i, x_i|\Theta)   - F(u, j, x_j|\Theta) $ in Equation (\ref{Equ:attack}) can be achieved by maximizing the BPR loss function $\mathcal{L}_{BPR}$ as defined in Equation (\ref{Equ:bpr}), namely, ${\operatorname{argmax}}_{\Theta}  \{ - \sum_{(u,i,j) \in \mathcal{D}_s} \ln{\sigma (r_{ui}-r_{uj})} + \lambda \| \Theta \|^2 \}$. Figure \ref{fig:attack} illustrates the procedure of generating adversary examples in attacks to visually-aware recommender systems, and different attack methods, such as FGSM and PGD, can be applied.

\begin{itemize}

\item {\bf \noindent Fast Gradient Sign Method (FGSM)} \cite{goodfellow2014explaining} has the advantage of generating adversarial perturbations quickly over other methods. It needs only one step to generate the attacked image. Given a clean input image $x$, the adversarial example can be computed by a local perturbation computed by the fast gradient sign method.
\begin{equation} \label{eq:fgsm}
    x^{*} = x + \epsilon ~ sign \left( \nabla_{x} \mathcal{L}_{BPR}  \right),
\end{equation}
where $\epsilon$ corresponds to the magnitude of adversarial signals, $sign(\cdot)$ is the sign function and $\nabla_{x} \mathcal{L}_{BPR}$ is the gradient of the attack loss function. 

\item {\bf \noindent Projected Gradient Descent (PGD)} \cite{kurakin2016adversarial} is an iterative version of FGSM. The attack algorithm iterates FGSM with a smaller step size. After each completed perturbation step, the intermediate attacked image is clipped to a $\epsilon-$neighborhood of the original image $x$. 
\end{itemize}

Our goal is then to design an effective defense strategy that can simultaneously (1) mitigate the impacts of adversarial attacks on recommendation performance and (2) detect adversarial attacks. 
We assume that we have the user-item interaction data $\mathcal{D}=\{\langle u, i, r_{ui} \rangle\}$, the images associated with the items ${\mathcal{X}}=\{x_i\}_{i\in \mathcal{I}}$, and the perturbed images ${\mathcal{X}}^*=\{x_i^*\}_{i\in \mathcal{I}}$ generated according to Equation (\ref{Equ:attack}). 
Given data $\{\mathcal{D}, \mathcal{X}, {\mathcal{X}}^*\}$, our objective is to build a denoising network to reconstruct image items for boosting the robustness of the recommendation, 
and a detection network to detect adversarial samples.
To our knowledge, this formulation of joint \emph{robust model construction} and \emph{attack detection} for VARS has not been considered in the open literature.
\vspace{5pt}

{\bf \noindent Targeted Attacks to VARS}. Note that, as shown in Section \ref{Sec:targeted-attacks}, our proposed defense can  be extended to  targeted attacks to VARS. Unlike untargeted attacks, which aim to disrupt the system’s performance broadly, targeted attacks aim to   promote a target item  so that the target item is recommended to as many users as possible.
In the context of VARS, the vulnerabilities exposed by methods like TAaMR \cite{di2020taamr} and AIP \cite{liu2020adversarial}  illustrate how attackers can exploit the system’s reliance on visual data to achieve their aims. For instance, TAaMR focuses on deceiving a classifier within the recommendation process. This approach involves creating adversarial images that cause the system to misclassify a targeted item as a popular or high-ranking category, leveraging the classifier’s role in item identification. AIP, in contrast, targets the ranking mechanism directly, rather than the classifier. Formally, let $\mathbf{C}$ be a classifier such that $\mathbf{C}(x)=c$ for a clean image $x$, where $c$ represents the class of the item. Given a target class $t \neq c$, TAaMR seeks an adversarial example  $x^*$  by solving the following optimization problem: $\min_{\substack{d \leq \epsilon }} d(x, x^*) \ \text{such that} \ \mathbf{C}(x^*) = t$.
AIP exploits visually-aware recommenders by modifying the feature embeddings used by ranking algorithms to assess item similarity or relevance. This approach manipulates the representation of an item within the recommendation space, pushing it toward higher positions in ranked lists without altering its classification.

\subsection{Overview of Proposed Adversarial Image Denoising and Detection Framework}

\begin{figure}[!t]
\centering
\includegraphics[width=1.0\textwidth]{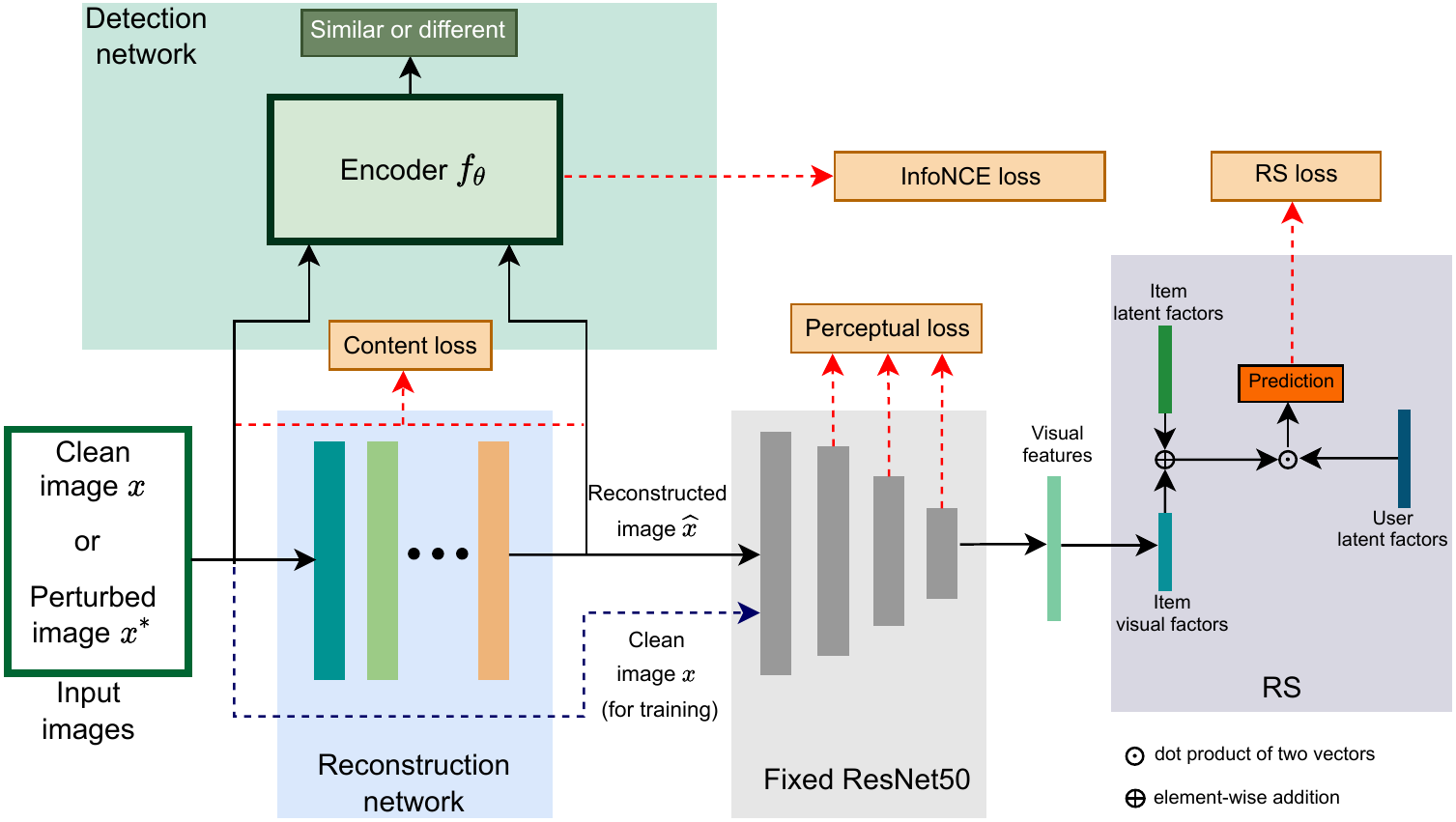}
\caption{The pipeline of our proposed framework. The defense architecture mainly consists of two parts:  (1) the denoising network to  reconstruct item images whether the input images have been attacked or not to  boost the recommendation robustness. Note that pairs of clean and perturbed images $\langle x, x^* \rangle$ are needed as input  to  train the denoising network; 
and (2) the detection network for detecting adversarial samples. Other parts in this figure are basically the same as the components of the VBPR model, which includes a pre-trained network (ResNet50 in this work) for image feature extraction and a latent factor  model for user preference prediction. \label{fig:pipeline}}
\end{figure}

To secure visually-aware recommendation systems (VARS) from adversarial attacks, we propose a framework that takes into account the robustness of the model to adversarial attack and the detection of adversarial examples. Meanwhile, our framework is designed to be used as a filter and detector prior to the recommendation system model. Therefore, our model can be trained first and then used as a pre-processing step without affecting the architecture and parameters of the current recommender system.  Figure \ref{fig:pipeline} shows the entire pipeline of our framework, which can be divided into three parts: the {\em denoising network} to remove adversarial signals, the {\em detection network}  to detect adversarial examples,  and the {\em recommendation system} (RS) model to predict the final preference scores of the items for each user. 

For the denoising part, we first sample the inputs $(u,i,j)$ from the training set 
$\mathcal{D}_{s}=\left\{(u, i, j) \right\},$
where the triplet $(u,i,j)$ indicates that the user $u$ prefers item $i$ over item $j$. 
Beyond the user-item interaction records, we have item images, clear images ${\mathcal{X}}=\{x_i\}_{i\in \mathcal{I}}$ or perturbed images ${\mathcal{X}}^*=\{x_i^*\}_{i\in \mathcal{I}}$, associated with the items.
In our work, the triplet $(u,i,j)$ is used as input of the recommendation model, and the corresponding image of the items is input of our proposed defense model. We assume that the input images can be clean (${\mathcal{X}}=\{x_i\}_{i\in \mathcal{I}}$) or perturbed (${\mathcal{X}}^*=\{x_i^*\}_{i\in \mathcal{I}}$) via adversarial perturbation. 
Our objective is to denoise/remove the adversarial signals while preserving the clean images. 
To achieve this, we construct a denoising network $T(\cdot)$, a neural network that maps both clean and perturbed images to denoised versions. Specifically, for a clean image $x$, the network generates $\hat{x} = T(x)$, and for a perturbed image $x^*$, it produces $\hat{x} = T(x^*)$. To train this network, we use pairs of clean and perturbed images $\langle x_i, x_i^* \rangle_{i \in \mathcal{I}}$ as input and apply a perceptual loss function, encouraging the denoised image $\hat{x}_i = T(x_i^*)$ to closely match the corresponding clean image $x_i$.

Once trained, this denoising network is employed to reconstruct item images, regardless of whether they have been attacked. The reconstructed images are expected to be as close as possible to clean images, so that the visual features extracted by downstream models (e.g., ResNet50) are not adversely affected by adversarial perturbations in the recommendation process.

For the detection part, based on the denoising network, we compare the denoised image with the input image. If they are similar, the input image is originally clean; if they are different, the input image might have experienced adversarial perturbation. Then the question boils down to a reliable evaluation of the similarity or differences between two images. Along this line of reasoning, we turn to metric learning \cite{koestinger2012large}, which aims to automatically construct domain-specific distance metrics from the training dataset. We then use the learned distance metric for other tasks, such as adversarial detection. In our case, the distance metric is used as a judgment of similarity; our detection network is based on the distance obtained by metric learning. 

A notable feature of the proposed system is the {\em joint training} of the detection and denoising networks. Note that the objectives of denoising and detection are mutually beneficial to each other. Denoising can help detection by filtering out adversarial attack signals; on the other hand, detection can facilitate denoising by pushing clean images away from noisy ones. Joint optimization of the denoising and detection modules in an end-to-end manner allows them to interact with each other for improved generalization performance, as will be experimentally verified later. To our knowledge, such end-to-end optimization of detection and denoising modules has not been proposed before. 
In the next three sections, we will first elaborate on the denoising network in Sec. \ref{sub:recon} and the detection network in Sec. \ref{sub:detect}. Then we present the total loss function for joint denoising and detection in Sec. \ref{sec:loss}.

\subsection{Image Denoising based on Residual and Transformer Blocks} \label{sub:recon}

The goal of image denoising within recommendation systems is to produce clean images by mitigating adversarial perturbations that could otherwise hinder recommendation performance. VARS models, such as the VBPR model, typically use convolutional neural networks (CNNs) to extract image features from item images before integrating these features into recommendation models like the widely used BPR model. The observed degradation in recommendation performance arises because adversarial signals can activate semantically irrelevant regions within the intermediate feature maps, disrupting the task-specific features that CNNs (e.g., ResNet50) extract from item images \cite{xie2019feature}. Although these perturbations may be imperceptible to humans, they are amplified in the higher layers of CNNs, which amplifies misleading characteristics that ultimately compromise recommendation accuracy.

\begin{figure}[tb]
\centering
{\includegraphics[width=1.0\textwidth]{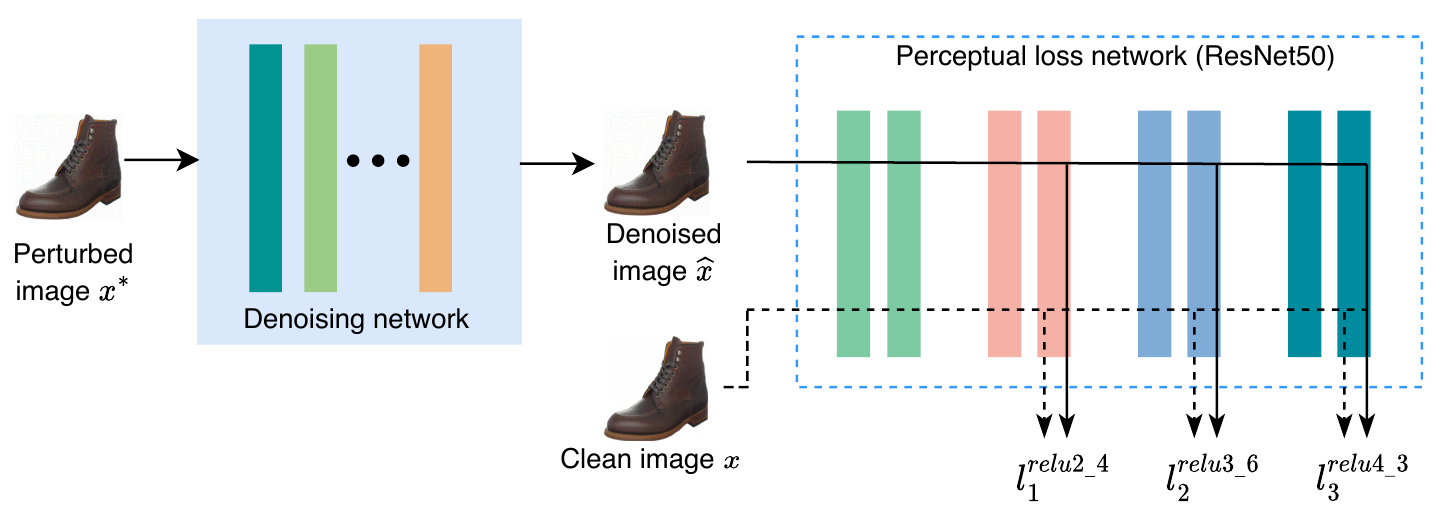}}
\caption{Illustration of the denoising network. The denoising network transforms a perturbed image $x^*$ into a denoised image $\hat{x}=T(x^*)$. Perceptual loss  that measures the feature differences
between the clean image $x$ and denoised image $\hat{x}$ at different intermediate feature maps is used to supervise the training of the denoising network.}
\label{fig:perceptual}
\end{figure}

Now the problem of defense against adversarial attacks boils down to how to denoise the intermediate feature maps generated by the layers of the neural network used for image extraction. 
Inspired by \cite{xie2019feature, zhang2021defense}, we devised a denoising network to reconstruct the images from their perturbed images to increase the robustness of the recommendation. The denoising network $T(\cdot)$ is a neural network that transforms a perturbed image $x^*$ into a denoised image $\hat{x}$ through $\hat{x}=T(x^*)$. As shown in Figure \ref{fig:perceptual}, to train the denoising network, we take pairs of clean and perturbed images $\langle x, x^* \rangle$ as input and apply perceptual loss to ensure that the denoised image $\hat{x}=T(x^*)$ is similar to the corresponding clean image $x$.

\begin{figure}[tb]
\centering
{\includegraphics[width=1.0\textwidth]{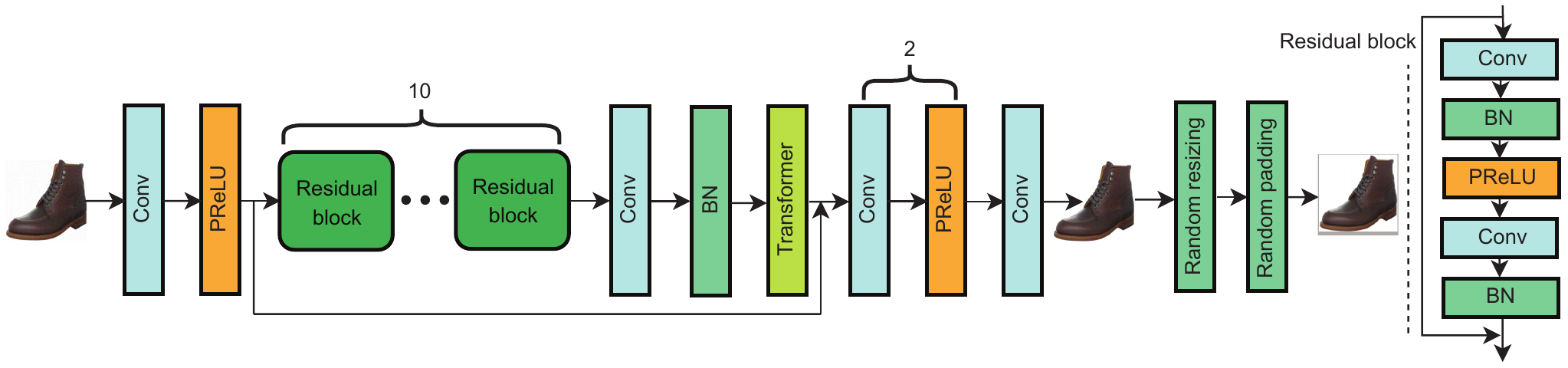}}
\caption{The denoising network architecture. It primarily consists of a series of 10 residual blocks (dark green) and transform blocks (light green) along with the basic convolution operations with kernel size 3 (except for the first and last one whose kernel size is 9). For the activation function, we have adopted both parametric ReLU (PReLU) and Batch Normalization (BN). Note that we can append one random layer that contains random resizing and padding after the final layer at inference time. }
\label{fig:recon}
\end{figure}

The primary components of the denoising network are shown in Figure \ref{fig:recon}. The residual blocks are the core components of the image denoising network. 
Because the goal of the denoising network is to remove adversarial perturbations,  a residual block is a good choice as it excels in learning the difference (or the residual) between input and output. Residual blocks are able to keep features for the clean image
that focus primarily on semantically informative content in the image, and remove feature maps for the adversarial image that 
are activated across semantically irrelevant regions. 
Furthermore, recent advances in visual recognition (\eg bottleneck transformer \cite{srinivas2021bottleneck}) have shown that it is better to replace spatial convolutions with $global$ self-attention in the final bottleneck block of the denoising network for the task of visual recognition. Based on the observation that adversarial attacks are local perturbations, we advocate the inclusion of a global vision transformer block in image denoising. The entire network consists of nine residual blocks followed by a transformer block and some separated convolution operations. The first and last convolutions are equipped with a kernel size of 9, and the others with 3. To generate clean denoised images, the entire pipeline does not involve any downsampling operators. Following \cite{xie2018mitigating}, in this work, random operations at inference time are also used to mitigate the adversarial effect. Specifically, two random operations are imposed on the denoised images. The denoised image of size $224 \times 224$ is first randomly resized to a smaller image $n\times n, n\in [212, 224]$, and then the resized image is randomly padded with zeros to the size of $224 \times 224$ with zeros.

We use the perceptual loss of the image \cite{johnson2016perceptual} to supervise the task of denoising feature maps. 
As shown in Figure \ref{fig:perceptual}, the perceptual loss function is calculated by comparing high-level differences based on intermediate features extracted from pre-trained networks. Note that the perturbation signals will be amplified in the high-level features
of the CNN network. We expect the intermediate features derived from the denoised images to be as close to the features extracted from clean images as possible. Specifically, given a pre-trained neural network $\phi$, let $\phi_{l}(x)$ be the features of the $l-$th convolution layer of the network $\phi$, $x$ and let $\hat{x}$ be the input image and the denoised image. The perceptual loss in the $l-$th convolution layer can be defined as
\begin{equation} \label{eq:perc}
    \mathcal{L}_{perc}^l = \sum_{x \in \mathcal{X}}\mathrm{Dist} \left( \phi_{l}(x), \phi_{l}(\hat{x})  \right),
\end{equation}
where $\mathrm{Dist}(\cdot)$ denotes the $L_2$ distance function. As studied in \cite{deldjoo2021study}, ResNet50 has shown quantitatively and qualitatively to produce the best recommended products. Therefore, we use ResNet50 \cite{he2015deep} as our pre-trained network for the computation of perceptual loss. 
In particular, we use the outputs of stages 2, 3, and 4 of the ResNet50, denoted as  $l_1^{\mathrm{relu2\_4}}$, $l_2^{\mathrm{relu3\_6}}$ and $l_3^{\mathrm{relu4\_3}}$, to retrieve the high-level features to compute the distance differences. Then the perceptual loss $\mathcal{L}_{perc}$ is the sum of the perceptual losses in the layers of 
$l_1^{\mathrm{relu2\_4}}$, $l_2^{\mathrm{relu3\_6}}$ and $l_3^{\mathrm{relu4\_3}}$. 

In addition to perceptual loss, in the denoising network, we also use content loss to supervise the image generation task. The content loss is defined as the $L_{2}$ of the different between the input image $x$ and the constructed image $\hat{x}$, 
\begin{equation} \label{eq:pixel}
    \mathcal{L}_{pix} = \sum_{x \in \mathcal{X}} \|x - \hat{x}\|^2_{2}.
\end{equation}
Content loss is employed to obtain a slightly blurred image (i.e., filter out adversarial perturbation), while perceptual loss can also help the denoising model preserve rich details in an image.

\subsection{Attack Detection Based on Contrastive Learning} \label{sub:detect}

Beyond building a more robust model, attack detection is also an important strategy to defend against attacks on recommendation systems. In particular, the purpose of attack detection in this paper is to distinguish an innocent image from a malicious image with adversarial perturbations (adversarial example), which can degrade the performance of the recommendation model. We propose a contrastive learning-based approach to the detection of adversarial examples.  Contrastive learning \cite{chen2020contrastive} aims to learn representations of images that push away dissimilar examples and bring similar examples closer. As shown in Figure \ref{fig:pipeline}, we will denoise the image by $\hat{x}=T(x^*)$ when the item image is perturbed. We can detect the perturbed image $x^*$ if its representation is away from the representation of the denoised image $\hat{x}$. Similarly, when the item image is innocent, namely, the input is a clean image $x$, we would expect its representation to be close to that of the denoised image $\hat{x}=T(x)$.

To this end, we build the detection network $f_{\theta}(\cdot)$ which includes two parts: a neural network encoder $g(\cdot)$ to extract image features and a small projection head $h(\cdot)$ to project the extracted image space to a common embedding space for attack detection purpose. In particular, we use ResNet50 for $g(\cdot)$ and a  multi-layer perceptron (MLP) for $h(\cdot)$. 
Without causing ambiguity, let $x$ be the input image for the detection network\footnote{Note that the input image for the detection network can be a perturbed image $x^*$ or a clean image $x$ along with the denoised image $\hat{x}=T(x^*)$  or $\hat{x}=T(x)$ respectively.}, then we have its image representation in the embedding space as $\mathbf{z} = f_{\theta}(x)$. Specifically, we have: 
\begin{equation}\label{Equ:feaRepr}
\begin{split}
\mathbf{s}& =g(x)=\mathrm{ResNet50}(x)\\
\mathbf{z}& =h(\mathbf{s})=W^{(2)}\sigma(W^{(1)}\mathbf{s}), 
\end{split}
\end{equation}
where $\mathbf{s} \in \mathbb{R}^{2048}$ is the result of the average pooling layer in ResNet, $\sigma$ indicates the nonlinear activation function of ReLU and $W^{(1)}$, $W^{(2)}$ are the weights in the MLP. The final output $\mathbf{z} \in \mathbb{R}^{128}$ is the feature vector in the embedding space, and will be used to calculate the similarity to the other. The feature vectors of clean images will be pulled close in the embedding space, while those of adversarial images will be pushed away.

\begin{figure}[tbh]
{\includegraphics[width=0.8\textwidth]{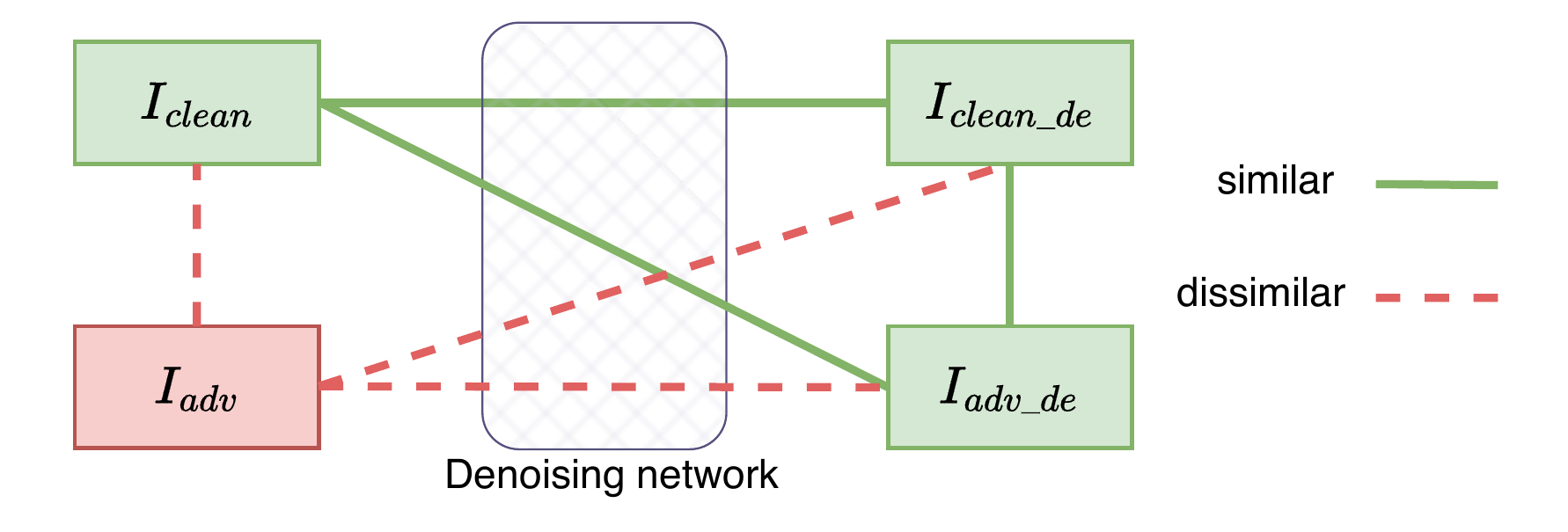}}
\caption{Positive and negative pairs for training the detection network via contrastive learning. The pair connected by a green line means positive pair (expected to be pulled close), and the pair connected by the red dot line means negative pair (expected to be pushed away). }
\label{fig:pos_neg_pairs}
\end{figure}

We then train the encoder with contrastive learning, which encourages the clustering of different classes of samples around their centroids.
Specifically, for each item with a corresponding image $x$, we first construct its positive and negative pairs as training data samples for encoder training $f_{\theta}$ based on inputs $I_{clean},~ I_{adv}$ and the corresponding outputs $I_{clean\_de},~ I_{adv\_de}$ from the denoising network. The feature vectors of the pairs are then placed close to each other in the embedding space if two samples in the pair are similar; otherwise, the feature vectors of dissimilar samples in the pair are separated by a large distance from each other. Figure \ref{fig:pos_neg_pairs} shows the construction of positive and negative pairs. Ideally, we assume that our denoising network will produce clean images without adversarial perturbations for both natural and adversarial examples. For each example $I_{clean}$, we will have a corresponding adversarial version $I_{adv}$. When our denoising network is completed, the denoised versions $I_{clean\_de}$ and $I_{adv\_de}$ will be generated. Based on the above observation, we expect that different distances will be learned between similar and dissimilar pairs. Positive and negative pair sets can be constructed as follows.

\begin{equation}
S_{pos} = \left \{
\begin{matrix}
(I_{clean}, I_{clean\_de}), \\
(I_{clean}, I_{adv\_de}), \\
(I_{clean\_de}, I_{adv\_de})
\end{matrix}
\right \} \ S_{neg} = \left \{
\begin{matrix}
(I_{adv}, I_{clean}), \\
(I_{adv}, I_{adv\_de}), \\
(I_{clean\_de}, I_{adv})
\end{matrix}
\right \}
\end{equation}

The similarity of two samples in the pair can be measured by cosine similarity. Let $sim(\textbf{u}, \textbf{v}) = \textbf{u}^\top \textbf{v}/ \left \| \textbf{u} \right \| \left \| \textbf{v} \right \| $ denote cosine similarity. For training the encoder $f_{\theta}$, we propose to minimize the following contrastive loss similar to the InfoNCE loss \cite{oord2018representation}:

\begin{equation}
    \mathcal{L}(x) = -\log \frac{\sum\limits_{x_i, x_j \in S_{pos}} \exp (sim (f_{\theta}(x_i),f_{\theta}(x_j))/\tau)}
    {\sum\limits_{x_{i},x_{j} \in S} \exp (sim (f_{\theta}(x_{i}),f_{\theta}(x_{j}))/\tau) },
\end{equation}
where $x_{i}, x_{j}$ are derived from item image $x$, $S=S_{pos}\cup S_{neg}$, and $\tau$ is a temperature hyperparameter \cite{wu2018unsupervised}. Given a dataset $X$, the total contrastive loss will be calculated as 
\begin{equation}
    \mathcal{L}_{contr}= \underset{x \sim X}{\mathbb{E}}\left[ \mathcal{L}({x}) \right].
\end{equation}

After obtaining a well-trained feature encoder $f_{\theta}$, its output can be used to calculate the similarity between two samples. Let $I_{in}$ (perturbed image $x^*$ or clean image $x$) and $I_{out}$ (denoised image $\hat{x}=T(x^*)$  or $\hat{x}=T(x)$) be the input and output of our denoising network. The feature vectors can be obtained by $z_{in}=f_{\theta}(I_{in})$ and $z_{out}=f_{\theta}(I_{out})$. Then we use the Euclidean distance between $z_{in}$ and $z_{out}$ as the dissimilarity score. A high score indicates that $I_{in}$ and $I_{out}$ are dissimilar, which implies that $I_{in}$ is an adversarial example because $I_{out}$ is purified after going through our denoising network. In contrast, a small score indicates that $I_{in}$ and $I_{out}$ are similar and both are likely to be clean examples. 
The optimal threshold for the decision boundary can be empirically determined by examining  the distribution of the distances between pairs of two classes (clean vs. noisy)  learned by detection network. For example,  as illustrated in Figure \ref{fig:distribution}, we set the threshold as $0.2$ in our experiments.

\subsection{Total Loss Functions for Joint Denoising and Detection}
\label{sec:loss}

As shown in Figure \ref{fig:pipeline}, there are four kinds of loss function in total in our framework. The detection network is optimized by contrastive loss, InfoNCE loss \cite{oord2018representation} in our case, which helps the detection network to learn an embedding function that maps the image $x$ to the feature $z$. This embedding will pull the two clean samples close and push a clean sample and an adversarial sample away in the embedding space. For image quality enhancement and removal of adversarial signals, we equip the denoising network with content loss and perceptual loss. For the recommender system, we use the same loss function as VBPR because we use VBPR as our recommendation model in the backend. The loss function of VBPR is defined as \cite{he2016vbpr}:
\begin{equation} \label{eq:bpropt}
    \mathcal{L}_{rs} = \underset{\Theta}{\operatorname{argmin}} \sum_{(u, i, j) \in \mathcal{D}_{s}}-\ln \sigma\left(\widehat{r}_{u, i}-\widehat{r}_{u, j}\right)+\lambda_{\Theta}\|\Theta\|^{2},
\end{equation}
where $(u, i, j)$ is the sampled pair of items of user $u$, $\widehat{r}_{u, i}$ is the preference score of user $u$ for item $i$, $\Theta$ represents all parameters of the model, $\lambda_{\Theta}$ is the weight of the regularization term and $\sigma$ is the Sigmoid function. To avoid overfitting \cite{tkde20amr}, we define the user preference score $u$ for item $i$ as $ \widehat{r}_{u, i}=\gamma_{u}^{\top}\left(\gamma_{i}+E f_{i}\right), $
where $\gamma_{u}$ and $\gamma_{i}$ are vectors $K-$ dimensional (e.g., $K=64$) that represent the latent factors of the user $u$ and the item $i$, respectively. Furthermore, $f_{i} \in \mathbb{R}^{2048}$ is a feature vector of the item $i$ extracted by a pre-trained neural network, and $E \in \mathbb{R}^{K \times 2048}$ is an embedding matrix that maps $f_{i}$ into a $K$ dimensional latent space. 

In summary, we add these loss functions together to jointly optimize our denoising and detection networks as
\begin{equation} \label{eq:loss}
    \mathcal{L} = \mathcal{L}_{pix} + \alpha \mathcal{L}_{perc} + \beta \mathcal{L}_{contr} + \xi  \mathcal{L}_{rs}, 
\end{equation}
where $\alpha, \beta, \xi $ are hyperparameters. It should be noted that the detection and denoising networks with their corresponding loss are optimized jointly. Based on our constructed multiple types of positive and negative pairs, the detection network can provide useful feedback signals to help the denoising network suppress adversarial perturbations and vice versa.

\subsection{Time Complexity Analysis} \label{sub:bigo}
We analyze the time complexity of our proposed framework in comparison with the baseline Visual Bayesian Personalized Ranking (VBPR) model, which we use as a foundation. The primary difference between our framework and VBPR lies in the additional operations involved in defense against adversarial attacks.

To more effectively convey the time complexity during training, let  $O_f$  represent the time complexity for forward propagation,  $O_b$  for backward propagation, and  $O_u$  for updating parameters. Additionally, let  $O_d$  denote the time complexity for adversarial image denoising and  $O_c$  for attack detection using contrastive learning. In VBPR, the training complexity consists of two  $O_f$ , two  $O_b$ , and one  $O_u$  for each sample pair due to the pairwise ranking loss, totaling  $2 \times O_f + 2 \times O_b + O_u$ . In our defense framework, the time complexity includes these operations along with the additional  $O_d$  for denoising and  $O_c$  for detecting adversarial perturbations. Thus, our time complexity for each sample pair becomes  $2 \times O_f + 2 \times O_b + O_u + O_d + O_c$ . Typically, these complexities are linearly correlated with $K$  and  $D$, where $K$ is the dimension of latent factors and $D$ is the dimension of visual feature vectors extracted by $\mathrm{ResNet50}$.   Therefore, for both VBPR and our defense framework, the time complexities are  $O(K + KD)$. 

\section{Experiments}
In this section, we report our experimental results on two real-world datasets to show the effectiveness of our proposed framework on improving the robustness to adversarial attacks and the detection of adversarial examples.
We first evaluate our defense framework under untargeted attack setting. Then we evaluate the effectiveness of our defense against targeted attacks in
Section \ref{Sec:targeted-attacks}.

\subsection{Datasets}

Our experiments are conducted on two real-world datasets: \texttt{Amazon Men} and \texttt{Amazon Fashion}, both of which are derived from the Amazon Web store \cite{mcauley2015image}. The original dataset\footnote{\url{http://jmcauley.ucsd.edu/data/amazon/}}  contains over 180 million relationships among almost 6 million objects, which are the result of recording the product recommendations of more than 20 million users. The visual features of these two datasets have been shown to provide meaningful information for recommendation\cite{he2016vbpr,grauman2020computer}. User review histories are viewed as implicit feedback, which can be used to sample the triplet $(u,i,j)$, each item paired with an image to extract visual features. 
For data preprocessing, we downloaded all available images from the provided repository. Each user’s rating was converted into a binary 0/1-valued interaction data, and users with fewer than five interactions $(|\mathcal{I}_{u}^{+}| < 5)$ were excluded to remove cold users. Following the protocol outlined in \cite{tkde20amr}, we applied a leave-one-out method to generate test sets, where one interaction per user was randomly selected for testing, with the remaining interactions used for training. The statistics of these datasets after preprocessing are summarized in Table \ref{tab:statistic}.

\begin{table}[t]
\caption{Statistics of the datasets we used in this work. }
  \label{tab:statistic}
	\begin{tabular}{lccc}
	    \hline
		\textbf{Dataset} & User\# & Item\# & Interaction\# \\\hline
		Amazon Men & 34,244 & 110,636 & 254,870 \\
		Amazon Fashion & 45,184 & 166,270 & 358,003 \\
		\hline
	\end{tabular}
\end{table}

\subsection{Evaluation Metrics}

Since our recommender system generates the top-$N$ list based on the computed preference scores, we use the Hit Ratio (HR) and Normalized Discounted Cumulative Gain (NDCG) \cite{tkde20amr} to measure the quality of the top-N lists generated. 
\begin{itemize}
    \item  \textbf{Hit Ratio (HR$@N$)}: given the top-$N$ recommendation list, we check if the groundtruth item is in the list. If yes, we mark 1 for this user and 0 otherwise.
    \item \textbf{Normalized Discounted Cumulative Gain (NDCG$@N$)}:  given the top-$N$ recommendation list, we consider the rank position of the groundtruth item in the list. The score decrease as the groundtruth`s rank goes lower.
\end{itemize}
The difference between these two metrics is that HR only considers whether the recommended item exists in the top-$N$ list, while NDCG further takes into account the position of the recommended item in the top-$N$ list. 
In our evaluation, calculating NDCG (Normalized Discounted Cumulative Gain) across all items for each user would be computationally extensive. To address this, we adopt a common sampling approach following \cite{di2020taamr, he2017neural}, where we randomly select 100 items that the user has not interacted with and include the ground-truth item among these. Specifically, for each user in the test set, we generate a list of these 100 unobserved items. We then compute preference scores for this list and rank the items accordingly. This strategy reduces computational load while still providing a reliable estimate of ranking quality. The NDCG is then calculated based on the rank of the ground-truth item in this list. We evaluate the metric for top-N lists with N set to 5, 10, and 20 to measure the effectiveness across different recommendation list lengths. For the detection part, we measure the detection performance according to the classification accuracy.

\subsection{Training Procedures}
As shown in Figure \ref{fig:pipeline}, the complete pipeline consists of three parts: the detection network, the denoising network, and the recommender system (RS). There  are three hyperparameters in the loss function as shown in Euqation (\ref{eq:loss}), 
in our experiments, we choose $\alpha=1.0$, $\beta=100.0$, and $\xi =0.1$. 
To optimize these three components in a more computationally efficient manner, we opt to process them separately before fine-tuning. Specifically, we first train our model on the Amazon Men dataset. We first spent about 100 epochs training the VBPR model, which can be used as the baseline RS model by both AMR \cite{tkde20amr} and our framework. The learning rate when training the baseline VBPR model is set to $1e-3$ in the first 80 epochs and decreases to $1e-4$ in the last 20 epochs. We train the VBPR model in the same way on two datasets to produce a well-trained model. After having a pre-trained VBPR model, we spent another five epochs fine-tuning the model with a learning rate of $1e-4$ in an adversarial training way as described in \cite{tkde20amr} to obtain the well-trained AMR model. 

Based on the pre-trained VBPR model, we then build our framework and train the detection network and denoising network, respectively. We first link our denoising network with the pre-trained VBPR model. To enforce the denoising network's learning how to recover clean images and remove adversarial signals from perturbed images, we have fixed the parameters of the VBPR model, i.e., there is no updating of the VBPR model when training the denoising network. During training, we do not activate the random layers of the denoising network because it does not help the training. We use them only in the testing phase to disturb the adversarial signals. To speed up denoising network training, we only use clean input and adversarial input generated by the FGSM method. Under this condition, it takes about 10 epochs with a learning rate of $1e-5$ to complete the training process. Then it will take another three epochs to retrain our model with clean input, FGSM-adversarial input, and PGD-adversarial input, respectively. The proportion of these three inputs is controlled to be close to $1:1:1$. We use the random number generator to determine which types of input should be used in one iteration. The learning rate is maintained at $1e-5$. 

The well-trained denoising network can be transferred to another dataset by fine-tuning for other small-number (e.g., 2-3) epochs. 
Finally, we train the detection network based on the well-trained denoising network. It takes about two epochs to complete the training. When training the detection network, the adversarial inputs we used are generated by the FGSM method with $\epsilon=16$ only. As shown in our analysis (Table \ref{tab:detection}) later, the accuracy of the detection network is robust to different choices of $\epsilon$ values. The proportion of clean and adversarial inputs is close to $1:1$. We note that the well-trained detection network can be effortlessly transferred to other datasets without extra fine-tuning.

\begin{table}[t]
\caption{Defense performance comparisons between different methods under various attack conditions on \textbf{Amazon Men} dataset.}
\centering
    \begin{tabular}{c|r|c|c|c|c|c|c}
        \hline 
        \multicolumn{2}{c|}{} & \multicolumn{3}{c|}{HR@N} & \multicolumn{3}{c}{NDCG@N} \\ 
        \cline{3-8}
        \multicolumn{2}{c|}{} & 5 & 10 & 20 & 5 & 10 & 20 \\
        \hline
        \multicolumn{8}{c}{VBPR} \\ \hline
        \multirow{4}{*}{Clean} &VBPR   &0.3122  &0.4261  &0.5564 &0.1123 &0.1290   &0.1486 \\
        &AMR    &0.3145  &0.4265 &0.5569 &0.1131  &0.1293   &0.1489 \\
        &OURS  &0.3118  &0.4260 &0.5563 &0.1121  &0.1290   &0.1484 \\
        &OURS-rand &0.3112 &0.4257 &0.5556 &0.1117 &0.1286 &0.1480 \\
        \hline
        \multirow{4}{*}{FGSM} &VBPR &0.0098	&0.0286	&0.2340	&0.0054	&0.0113	&0.0602 \\
        &AMR  &0.0092	&0.0262	&0.2200	&0.0051	&0.0105	&0.0565 \\
        &OURS &0.2819	&0.4103	&0.5989	&0.1919	&0.2332	&0.2800 \\
        &OURS-rand &0.2799	&0.4090	&0.5993	&0.1881	&0.2295	&0.2775 \\
        \hline
        \multirow{4}{*}{PGD} &VBPR  &0.0000  &0.0000  &0.0018  &0.0000  &0.0000  &0.0004 \\
        &AMR   &0.0000  &0.0000  &0.0019  &0.0000  &0.0000  &0.0005 \\
        &OURS  &0.0315  &0.0892  &0.4055  &0.0165  &0.0346  &0.1112 \\
        &OURS-rand  &0.1480  &0.2719  &0.5598  &0.0863  &0.1257  &0.1973 \\ \hline
        \multicolumn{8}{c}{DVBPR} \\ \hline
        \multirow{4}{*}{Clean} &DVBPR   &0.3473  &0.4740  &0.6189 &0.1249 &0.1435   &0.1653 \\
        &AMR    &0.3498  &0.4744 &0.6195 &0.1258  &0.1438   &0.1656 \\
        &OURS  &0.3468  &0.4739 &0.6188 &0.1247  &0.1435   &0.1651 \\
        &OURS-rand &0.3462 &0.4735 &0.6180 &0.1242 &0.1430 &0.1646 \\
        \hline
        \multirow{4}{*}{FGSM} &DVBPR &0.0109	&0.0318	&0.2603	&0.0060	&0.0126	&0.0670 \\
        &AMR  &0.0102	&0.0291	&0.2447	&0.0057	&0.0117	&0.0628 \\
        &OURS &0.3136	&0.4564	&0.6662	&0.2135	&0.2594	&0.3115 \\
        &OURS-rand &0.3113	&0.4550	&0.6666	&0.2092	&0.2553	&0.3087 \\
        \hline
        \multirow{4}{*}{PGD} &DVBPR  &0.0000  &0.0000  &0.0020  &0.0000  &0.0000  &0.0004 \\
        &AMR   &0.0000  &0.0000  &0.0021  &0.0000  &0.0000  &0.0006 \\
        &OURS  &0.0350  &0.0992  &0.4511  &0.0184  &0.0385  &0.1237 \\
        &OURS-rand  &0.1646  &0.3024  &0.6227  &0.0960  &0.1398  &0.2195 \\
        \hline 
    \end{tabular}
    \label{tab:metric-amazonmen}
\end{table}

\begin{table}[t]
\caption{Defense performance comparisons between different methods under various attack conditions on \textbf{Amazon Fashion} dataset.}
\centering
    \begin{tabular}{c|r|c|c|c|c|c|c}
        \hline
        \multicolumn{2}{c|}{} & \multicolumn{3}{c|}{HR@N} & \multicolumn{3}{c}{NDCG@N} \\ 
        \cline{3-8}
        \multicolumn{2}{c|}{} & 5 & 10 & 20 & 5 & 10 & 20 \\
        \hline
        \multicolumn{8}{c}{VBPR} \\ \hline
        \multirow{4}{*}{Clean} &VBPR &0.3387 &0.4677 &0.6058  &0.2328 &0.2745 &0.3094 \\
        &AMR &0.3389 &0.4680 &0.6061  &0.2338 &0.2754 &0.3104 \\
        &OURS &0.3376 &0.4675 &0.6058  &0.2326 &0.2744 &0.3093 \\
        &OURS-rand &0.3330 &0.4663 &0.6067  &0.2285 &0.2711 &0.3070 \\
        \hline
        \multirow{4}{*}{FGSM} &VBPR &0.0026 &0.0116 &0.1854 &0.0014 &0.0042 &0.0453 \\
        &AMR &0.0026 &0.0110 &0.1747 &0.0014 &0.0040 &0.0427 \\
        &OURS &0.3045 &0.4412 &0.6019 &0.2035 &0.2477 &0.2881 \\
        &OURS-rand &0.2948 &0.4329 &0.6025 &0.1945 &0.2392 &0.2822 \\
        \hline
        \multirow{4}{*}{PGD} &VBPR &0.0000 &0.0000 &0.0025 &0.0000 &0.0000 &0.0006 \\
        &AMR &0.0000 &0.0000 &0.0027 &0.0000 &0.0000 &0.0007 \\
        &OURS &0.0133 &0.0564 &0.3580 &0.0065 &0.0201 &0.0929 \\
        &OURS-rand &0.1269 &0.2653 &0.5564 &0.0700 &0.1144 &0.1866 \\ \hline 
        \multicolumn{8}{c}{DVBPR} \\ \hline
        \multirow{4}{*}{Clean} &DVBPR &0.3768 &0.5202 &0.6739  &0.2590 &0.3053 &0.3442 \\
        &AMR &0.3770 &0.5206 &0.6742  &0.2601 &0.3063 &0.3453 \\
        &OURS &0.3755 &0.5200 &0.6739  &0.2587 &0.3052 &0.3440 \\
        &OURS-rand &0.3704 &0.5187 &0.6749  &0.2542 &0.3016 &0.3415 \\
        \hline
        \multirow{4}{*}{FGSM} &DVBPR &0.0029 &0.0129 &0.2062 &0.0016 &0.0047 &0.0504 \\
        &AMR &0.0029 &0.0122 &0.1943 &0.0016 &0.0044 &0.0475 \\
        &OURS &0.3387 &0.4908 &0.6695 &0.2264 &0.2755 &0.3205 \\
        &OURS-rand &0.3279 &0.4815 &0.6702 &0.2164 &0.2661 &0.3139 \\
        \hline
        \multirow{4}{*}{PGD} &DVBPR &0.0000 &0.0000 &0.0028 &0.0000 &0.0000 &0.0007 \\
        &AMR &0.0000 &0.0000 &0.0030 &0.0000 &0.0000 &0.0008 \\
        &OURS &0.0148 &0.0627 &0.3982 &0.0072 &0.0224 &0.1033 \\
        &OURS-rand &0.1412 &0.2951 &0.6189 &0.0779 &0.1273 &0.2076 \\
        \hline
    \end{tabular}
    \label{tab:metric-amazonfashion}
\end{table}

\subsection{Defense Performance and Analysis}

Our denoising network first strives to generate clean images regardless of the input images, whether they are clean or adversarial. The denoised images will then be fed into the following pre-trained neural network, which extracts visual features to be used by the recommender system models for predicting user preferences. To gain a deeper understanding, we will analyze the defense performance of our denoising network from the following perspectives: {\em Recommendation performance} and {\em transferability study}. We also present the visual quality of the denoised images. 

\textbf{Recommendation performance evaluation}. Table \ref{tab:metric-amazonmen} and \ref{tab:metric-amazonfashion} shows the quantitative results of our defense model in terms of HR@N and NDCG@N metrics on Amazon Men and Amazon Fashion datasets \cite{he2015trirank}. 
To better illustrate the generalization capability of our approach, we compare our defense model with two baseline methods, VBPR\cite{he2016vbpr} and DVBPR\cite{kang2017visually}. The results indicate that the proposed denoising network does not degrade the performance of existing recommendation models when evaluated on clean data. Additionally, we include a comparison with AMR \cite{tkde20amr}, which employs adversarial training by introducing small, deliberate perturbations to the deep image feature representations extracted by a pre-trained network. During training, an adversary generates subtle perturbations to these feature vectors, simulating potential adversarial attacks. The model adapts its learned representations to be invariant to these perturbations, thereby improving its overall generalization and resilience against adversarial attacks.  However, to reduce computational overhead, AMR generates adversarial perturbations only on intermediate visual features extracted by CNNs. In contrast, our method applies adversarial noise at the pixel level across the entire image.
As expected, the performance of VBPR, DVBPR and AMR drop dramatically in the presence of FGSM and PGD attacks on the RS model. A plausible reason for the failure of AMR is that adversarial perturbations are added to the positive image $i$ and the negative image $j$ based on the input sample triplet $(u,i,j)$. Perturbations on two-item images can not only decrease the rank position of the positive item, but also increase the possibility of recommending a negative item. In contrast, our defense model appears to be more robust to adversarial attacks than the AMR baseline. The denoising network can partially remove adversarial signals from the input images and effectively defend the RS model from FGSM attacks. When faced with more aggressive PGD attacks, our model alone becomes inadequate; however, with random operations (marked by ``(rand)'')  including random resizing and padding of images, it still achieves decent performance in challenging PGD attacks. We conclude that our proposed denoising network is an effective defense strategy against adversarial attacks without altering the original recommendation model.

\begin{figure}[t]
\centering
{\includegraphics[width=1.00\textwidth]{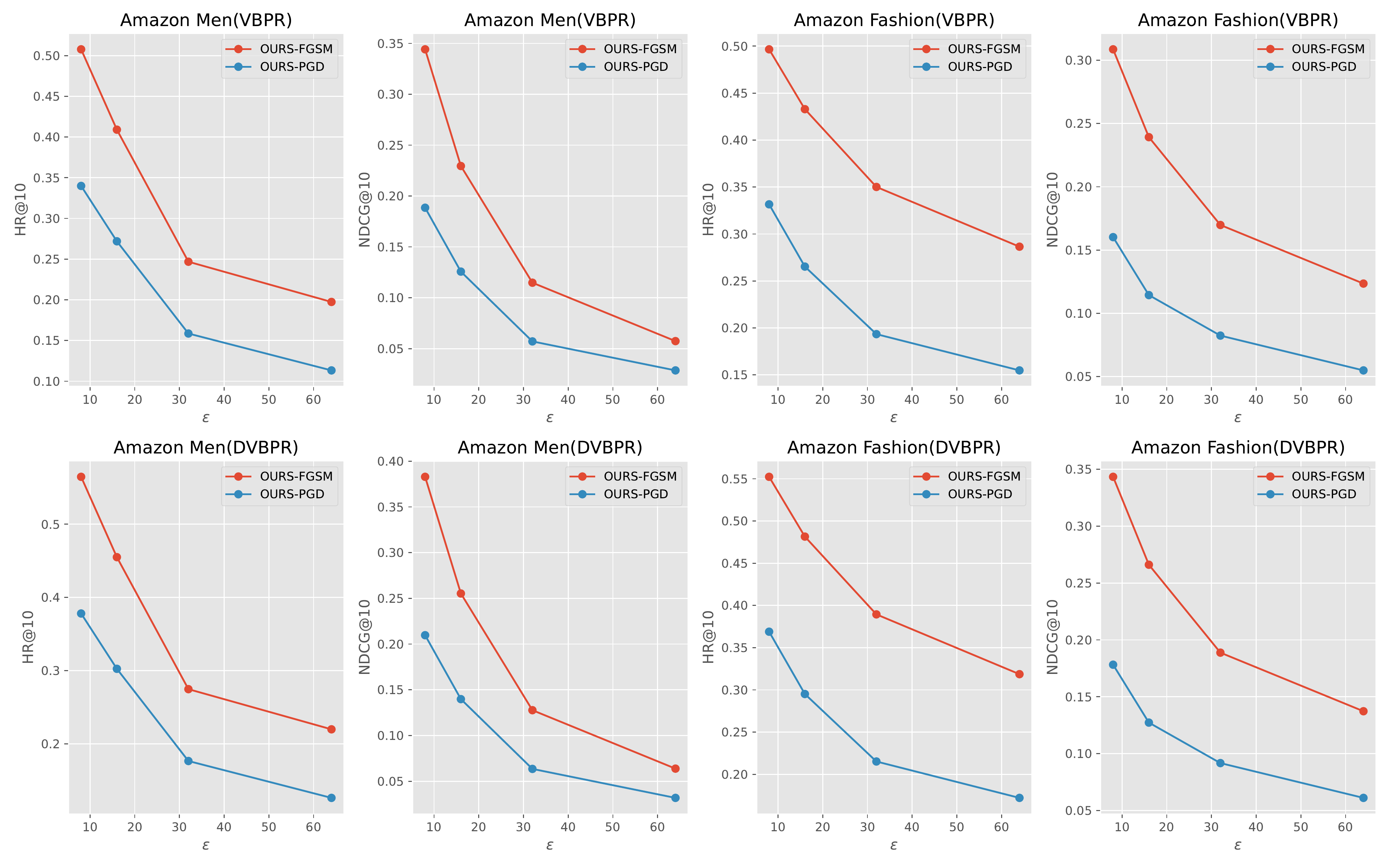}}
\caption{Performance of the denoising network w.r.t. different values of attack perturbation level $\epsilon$ (our model is trained  with $\epsilon=16$ only). }
\label{fig:epsilon}
\end{figure}

\textbf{Transferability study}. We want to demonstrate the transferability performance of our defense model when faced with attack signals with different strengths. Transferability is a desirable property for machine learning algorithms. As shown in the first column of Figure \ref{fig:epsilon}, attack signals with a smaller value $\epsilon=8$ do not greatly affect the performance of the RS model. However, for $\epsilon=32$ and $\epsilon=64$, the impact of adversarial perturbation on top-10 HR performance is more observable. Our denoising network manages to remove these adversarial perturbations while simultaneously improving visual quality. It has been empirically verified that the preservation of visual quality can be observed for different values $\epsilon$, although our denoising network is trained with $\epsilon=16$ only, which justifies the good transferability of our defense model.

\begin{figure}[tb]
\centering
{\includegraphics[width=0.8\textwidth]{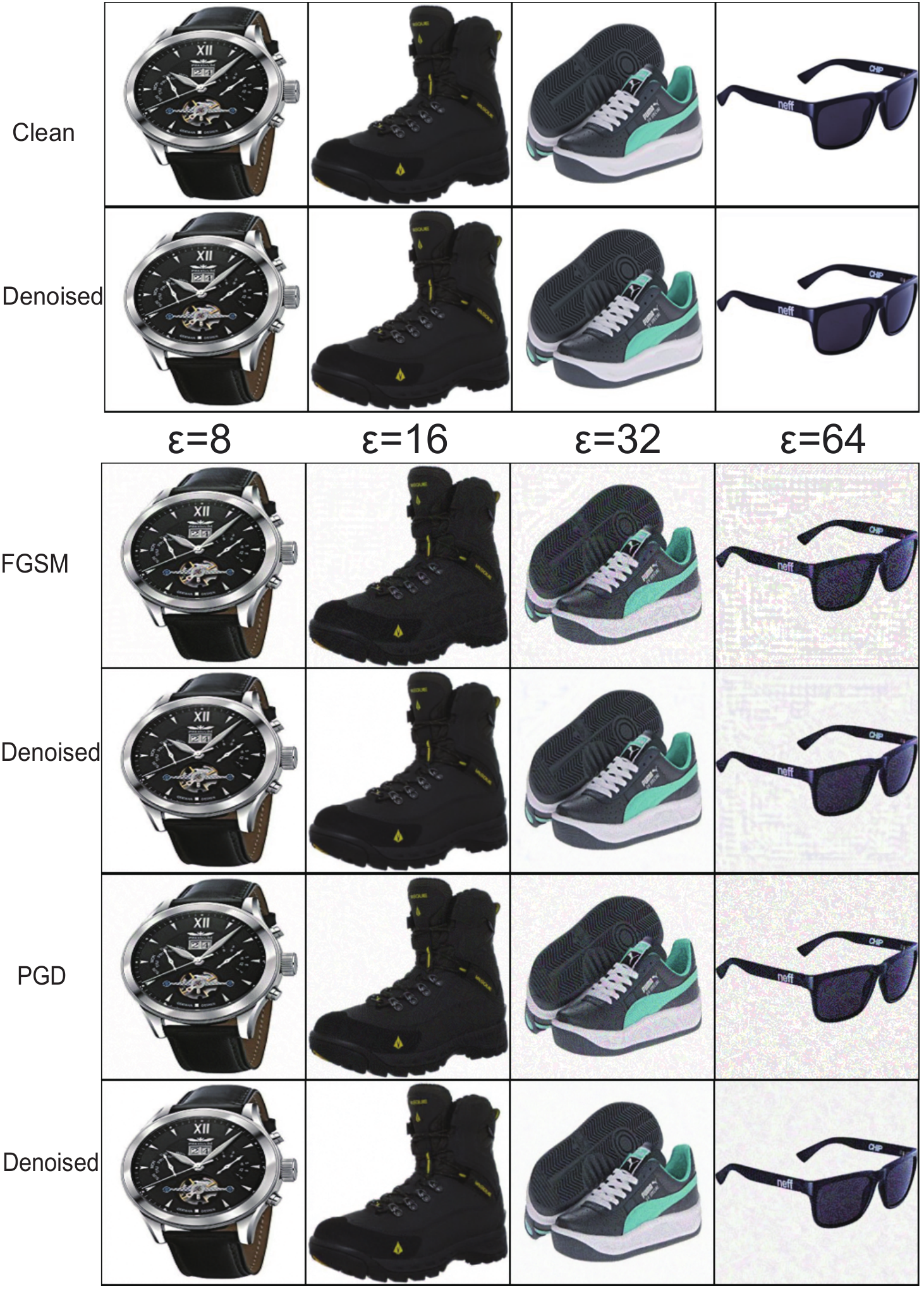}}
\caption{Denoised images from clean and adversarial images. FGSM and PGD are used to generate adversarial images. And we also investigate the effects of the magnitude of the attack perturbation magnitude $\epsilon$ on generated images. This figure is better viewed when zoomed in. }
\label{fig:vis}
\end{figure}

\textbf{Visual quality comparison}. Figure \ref{fig:vis} compares the images generated by our denoising network with different inputs (clean vs. adversarial). The figure shows the inputs and outputs of our denoising network under three circumstances and four different values of $\epsilon$. We manually adjusted this value to strike the best trade-off between attack efficiency and the imperceptibility of adversarial signals. In the presence of adversarial signals, there is a conflict between the objectives of preventing degradation of model performance and generating imperceptible perturbations. For example, if one wants the model performance to degrade gracefully,  the introduced attacked signals cannot be too large to become easily observable based on the visual appearance. Meanwhile, if the attacked signals remain imperceptible, they may be too weak to have an adversarial impact on the performance of the model. Based on the above analysis, we have handpicked $\epsilon=16$ so that our model can reach a good balance between attack efficiency and visual quality degradation. As shown in the second column of Figure \ref{fig:vis}, the visual quality of the denoised images from clean images is almost identical to that from adversarial input from FGSM and PGD attacks.

\subsection{Attack Detection Performance and Analysis}

 \textbf{Experimental Setup.} As shown in Figure \ref{fig:pipeline}, we can train our detection network based on the output images of the denoising network. We feed both the original input images and the corresponding denoised images into the detection network to obtain two feature vectors and compute the distance of these two feature vectors. By analyzing the distribution of the computed distance, we can learn the representations by maximizing feature consistency under differently augmented views. This way of separating adversarial examples from clean images is conceptually similar to the Adversarial-To-Standard (A2S) model in adversarial contrastive learning \cite{jiang2020robust}. More specifically, the dimension of the output of the feature vector of the detection network is 2048 in our current implementation. When evaluating our detection network, we have used 1000 items in the cold list as negative examples that have never been sampled to train the denoising network. Then we sample 1000 positive examples together with 1000 negative examples to obtain the testing set for evaluation. We control the percentage of adversarial examples in the original inputs to be around $50\%$.

\textbf{Baseline.} To the best of our knowledge, there was no prior work on detecting adversarial examples in visually-aware recommendation systems (VARS). 
We explored several methods for detecting adversarial samples \cite{tian2018detecting, liang2018detecting, wang2023addition}. It is important to note that all of these approaches rely on classifiers to identify adversarial examples. Specifically, given a well-trained classifier $\mathcal{C}$, an input $x$, and its transformed counterpart $\mathcal{T}(x)$, if $\mathcal{C}(x) \neq \mathcal{C}(\mathcal{T}(x))$, the input $x$ is treated as an adversarial example; otherwise, it is considered a clean example. However, our approach does not involve a classifier. Instead, it is based on regression, $r_{ui}=F(u, i, x_i|\Theta)$, meaning that previous detection methods are not directly comparable to ours. To facilitate a meaningful comparison, certain adjustments are required. For instance, we use a recent study \cite{wang2023addition} as a reference. We transform the problem of classification into a score difference between clean and noisy samples. Assuming that our VARS gives preference score $r_{ui}^*=F(u, i, x_i^*|\Theta)$ under attacks with perturbed image $x_i^*$, then we compute the score variance before and after attack $ |r_{ui} - r_{ui}^* | / r_{ui}$. 
If the score change reaches more than 20\%, we think the VARS model is affected by the attack. 
Based on this criterion, we compared our detection results to the detection results given by the model ADDITION in \cite{wang2023addition}. 
The ADDITION model leverages an adaptive two-stage process to detect adversarial examples. In the first stage, the noise injection module examines each input image to extract in-depth features and determines an optimal magnitude for additional Gaussian noise. This adaptive mechanism ensures that the injected noise is strong enough to mask any adversarial perturbations while preserving critical image details. In the second stage, the denoising module processes the noisy image to reduce the total noise—both the added Gaussian noise and any adversarial perturbations—to recover an image that closely approximates the original benign signal. We mainly implemented the noise injection module and noise reduction module in the work of ADDITION. The network architecture of the two modules follows the details given in the original paper.

\begin{figure}[tbh]
\centering
{\includegraphics[width=0.8\textwidth]{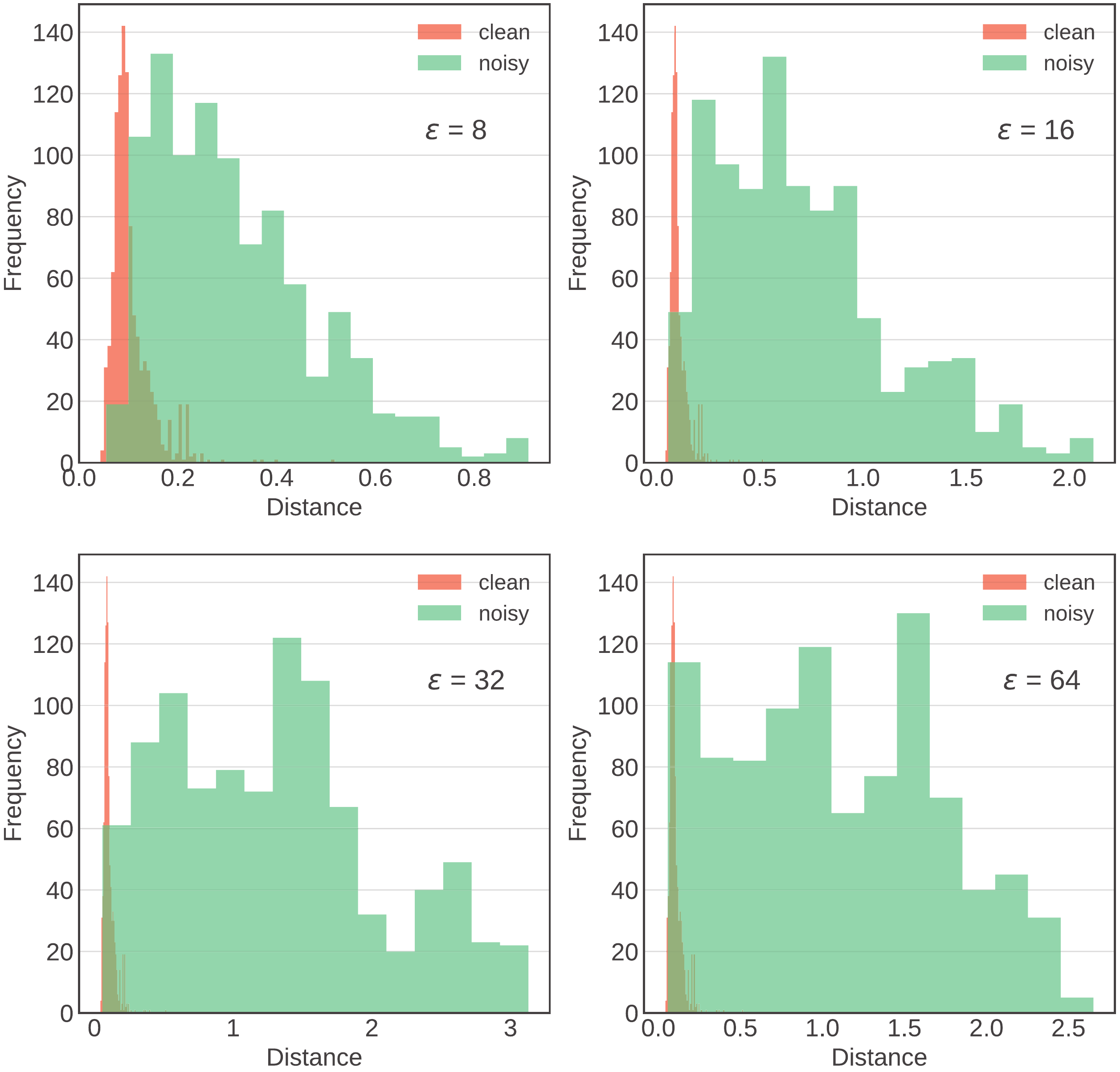}}
\caption{Distribution of two classes (clean vs. noisy) of distances learned by detection network with different $\epsilon$ values on Amazon Men dataset. }
\vspace{-10pt}
\label{fig:distribution}
\end{figure}

\begin{figure}[tbh]
\centering
{\includegraphics[width=0.6\textwidth]{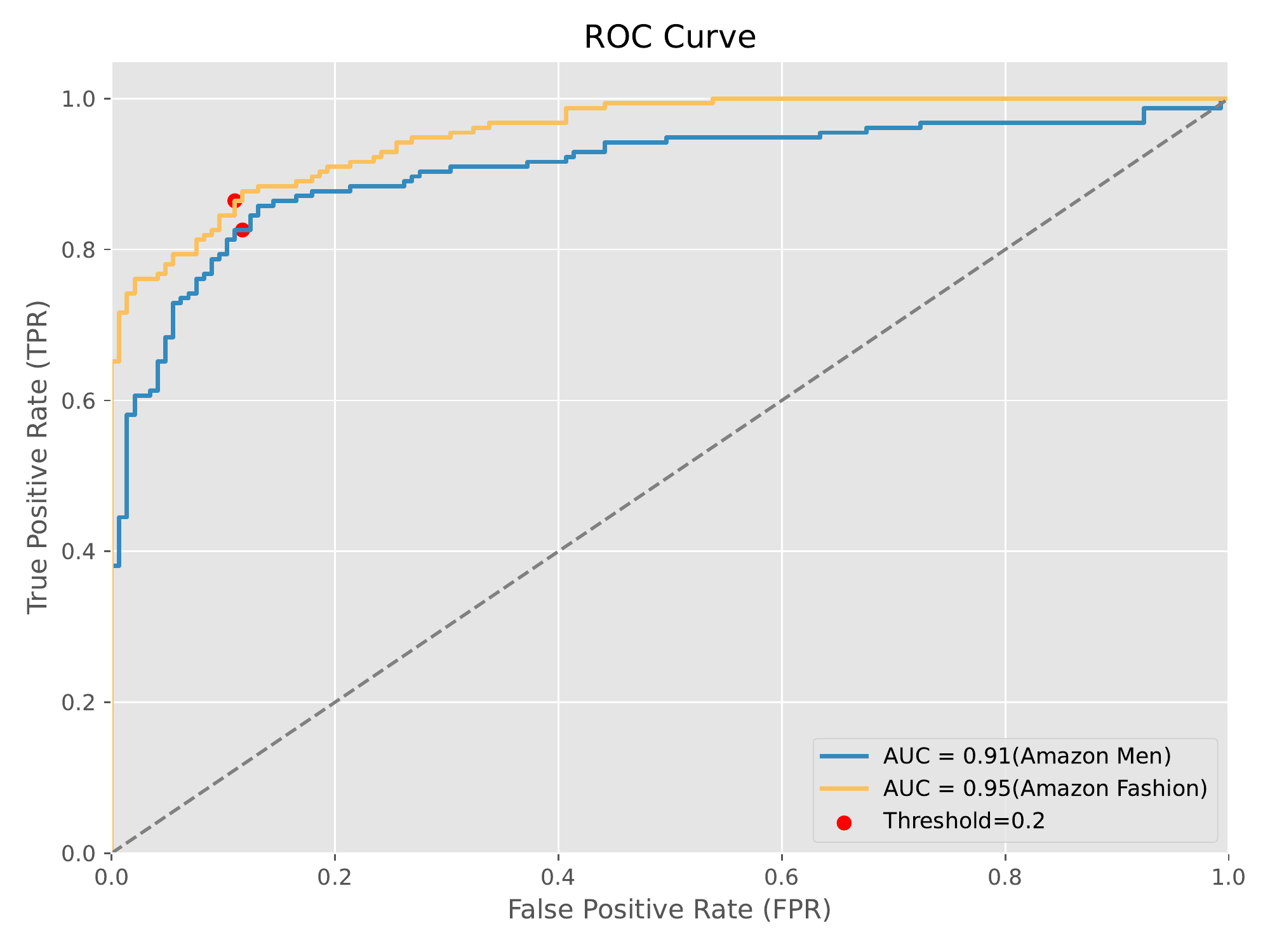}}
\vspace{-10pt}
\caption{ROC curves of attack detection   on two datasets. The red points mark the TPR/FPR at a distance threshold of 0.2. These results are obtained using the VBPR backbone.}
\label{fig:roc}
\end{figure}

\textbf{Threshold for adversarial example detection.} We  empirically set the threshold for adversarial example detection by examining the distribution of the distances between two classes (clean vs. noisy)  learned by the detection
network. Figure \ref{fig:distribution} compares the distributions of the distance profiles learned by the detection network on the Amazon Men data set. It can be seen that after training, the detection network has successfully separated clean samples from adversarial ones in the 128-dimensional feature space extracted. We can calculate appropriate distance thresholds for binary classification (noisy vs. clean). In our implementation, we use a distance threshold of $0.2$ to classify the test set into noisy and clean samples.  The threshold is further confirmed by the operating characteristic (ROC) curve, which displays model performance (\eg true positive rate (TPR) and false positive rate (FPR))  across different thresholds.   As shown in Figure \ref{fig:roc}, $0.2$ is selected as the threshold based on the balance it provides between the True Postive Rate (TPR) and the False Positive Rate (FPR). The red points in the figure mark the TPR/FPR values at this distance threshold.

\begin{table}[tb]
\centering
\caption{Detection accuracy under different perturbation level values. }
\label{tab:detection}
    {\begin{tabular}{rccccc}
    \hline  
    & perturbation level $\epsilon$   & 8     & 16    & 32    & 64 \\
    \hline
    \multirow{2}{*}{Amazon Fashion}  &ADDITION  &0.9211  &0.9325 &0.9410 &0.9572  \\
                                    &OURS  & 0.9523 & 0.9665 & 0.9636 & 0.9608 \\
                                   
    \hline
    \multirow{2}{*}{Amazon Men}     &ADDITION  &0.8419  &0.9044 &0.9158 &0.9415 \\
                                     &OURS  & 0.8826 & 0.9461 & 0.9660 & 0.9406 \\

    \hline 

    \end{tabular}}
\end{table}

\textbf{Results.} Table \ref{tab:detection} shows the high accuracy of the detection results in the two datasets, even with different values of $\epsilon$. In most cases, our detection network has high accuracy in distinguishing adversarial samples from clean ones.
We have to point out that the model ADDITION cannot distinguish large Gaussian noise from adversarial noise in the scenario of this work. When we sample noise from $\mathcal{N}(0, \sigma), \sigma \in \{8/255, 16/255, 32/255, 64/255\}$ and add the noise sampled to the clean images, ADDITION can only process noise with small $\sigma$. When it comes to $\sigma \in \{32/255, 64/255\}$, ADDITION cannot distinguish Gaussian noise and adversarial noise. For example, based on our experiments, when we sample noise from $\mathcal{N}(0, 64/255)$ and expand the test set of detection via adding Gaussian noise to clean images, our detection method can still have 95.23\% accuracy while ADDITION can only maintain 72.13\% accuracy on Amazon Fashion dataset.  In other words, ADDITION model cannot distinguish non-adversarial noise from adversarial noise. But our detection method can distinguish it accurately.

\begin{figure}
\centering
{\includegraphics[width=0.6\textwidth]{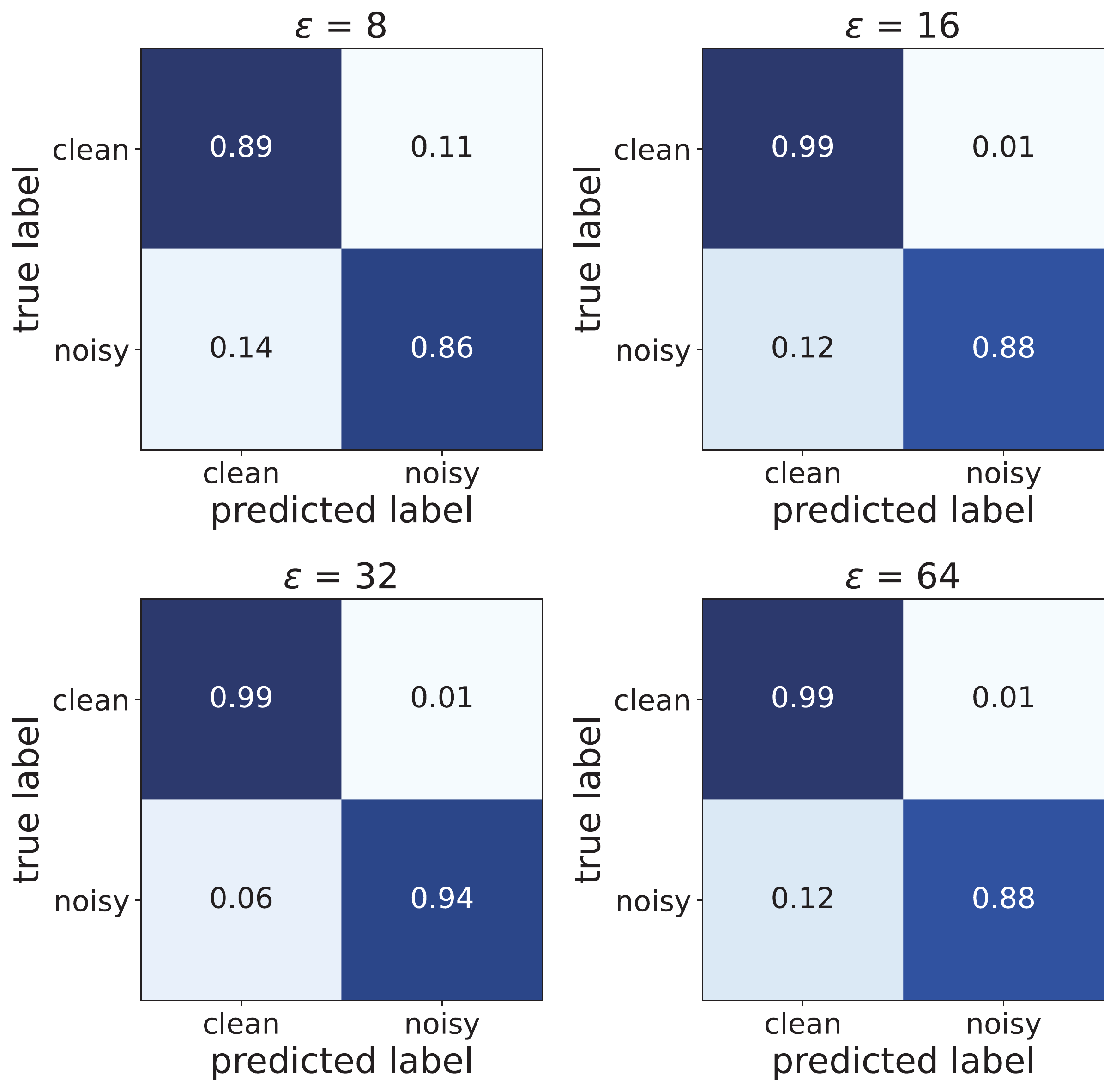}}
\caption{The detection confusion matrix with different values of the attack perturbation level $\epsilon$ in the Amazon Men dataset. }
\label{fig:confushion}
\end{figure}

 Figure \ref{fig:confushion} shows the confusion matrix of the detection results. We can see that our detection network can correctly divide the testing set into two parts: the adversarial set and the clean set in an ideal way. As mentioned in the last section, the defense performance of our denoising network decreases with increasing value $\epsilon$. However, this figure tells us that our detection network can still have a high detection accuracy in the case of large $\epsilon$. Although the performance of our detection network degrades slightly with decreasing $\epsilon$ values, our denoising network still performs well and maintains the performance of the original recommender systems. Through the experimental results, we find that our proposed detection network and denoising network complement each other in defense against adversarial attack. With a high $\epsilon$, the detection network predominates, while with a small $\epsilon$, the denoising network is more important.

\subsection{Interaction between Denoising and Detection Networks} \label{sec:recon_det}

The analysis in the previous subsection suggests that our detection network can have a high detection accuracy in the case of large $\epsilon$ values. Although the performance of our detection network decreases slightly with decreasing values of $\epsilon$, our denoising network still performs well and maintains the performance of the original recommender systems. Through empirical studies, we find that our proposed detection network and denoising network complement each other in defense against adversarial attack. With a high $\epsilon$, the detection network predominates because the performance of the denoising network degrades; while with a small $\epsilon$, the role played by the denoising network becomes more important because it is more challenging to detect the presence of subtle adversarial perturbations.

\begin{figure}
\centering
{\includegraphics[width=1.0\textwidth]{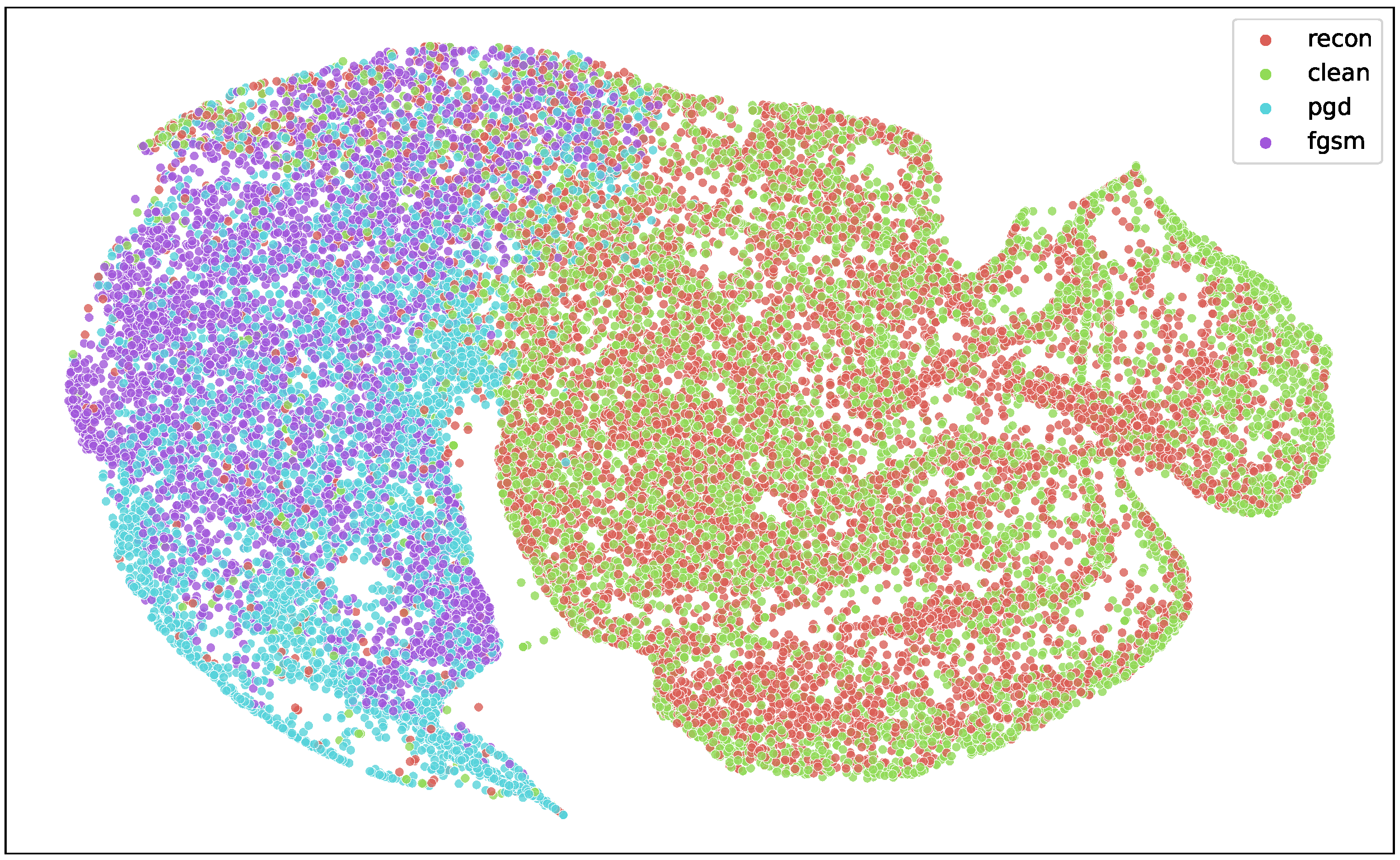}}
\caption{t-SNE visualization of four classes of images: clean images without perturbations (green), reconstructed images by our denoising network (red), and adversarial images with FGSM and PGD perturbations (blue and purple). }
\label{fig:tsne}
\end{figure}

To better illustrate the interaction between denoising and detection networks, we report the visualization result of t-SNE \cite{van2008visualizing} in Figure \ref{fig:tsne}. It can be clearly seen that the clusters of clean and denoised/reconstructed images (marked with green and red) are separated from those of adversarial images with FGSM and PGD perturbations (marked with blue and purple). This clear separation echoes the observation we made in Fig. \ref{fig:distribution}. When combined with the performance of the recommendations reported in Table \ref{tab:metric-amazonfashion} and \ref{tab:metric-amazonmen}, we conclude that the detection network and the denoising network mutually help each other by the joint training proposed with contrast loss.

\subsection{Ablation Study}
\label{Sec:ablation}

To further illustrate the different roles of certain internal components within this framework, we perform an ablation study, with a primary focus on the denoising module and perceptual loss. In this ablation study, we utilize the Amazon Men dataset and select PGD as the attack method, with a perturbation level of $\epsilon = 16$. The complete experimental results are presented in Table \ref{tbl:ablation}. For better comparison, we also include the results of our unmodified architecture under the same configuration settings.  

\emph{Detection network.}
The detection network serves as a crucial component of our framework. To demonstrate its importance, we conducted an ablation experiment by removing this module from our framework. Since the detection results depend on the detection network, it is not possible to report detection-related outcomes in its absence. As such, missing data in the corresponding table entries is indicated by NaN. The results, presented in Table \ref{tbl:ablation}, reveal a significant decline in the performance of the recommender system when the detection network is removed. This decline highlights the critical role the detection network plays in enhancing the denoising network’s ability to effectively suppress adversarial noise, underscoring its essential function within the framework.

\emph{Perceptual loss.}
The perceptual loss is a crucial loss in mitigating adversarial noise\cite{xie2019feature}. To evaluate its impact, we also conducted an ablation experiment by removing this loss function from the framework. The results show a dramatic degradation in both detection and recommendation performance without the perceptual loss, demonstrating its importance in strengthening the framework’s robustness against adversarial attack. These findings are consistent with prior research, such as \cite{xie2019feature, zhang2021defense}, further validating the effectiveness of perceptual loss in defending against adversarial attacks.

Through the ablation study, we can see that the denoising network and the   detection network play different roles in the defense against adversarial attack signals. They are mutually beneficial to each other in a unified framework. 

\begin{table}[t]
\caption{Impact of different components in the proposed framework.}
\label{tbl:ablation}
\centering
\begin{tabular}{c|c|c|c}
\hline 
\multicolumn{1}{c|}{metrics} & Accuracy & NDCG@10 &HR@10\\ \Xhline{3\arrayrulewidth}
unmodified architecture       & 0.9462 &0.1257 &0.2719 \\ \hline
remove detection \& contrastive loss  & NaN &0.0768 &0.1661 \\ \hline
remove perceptual loss      & 0.8521 &0.1031 &0.2230 \\ \hline 
\end{tabular}
\end{table}

\subsection{Defense on Targeted Attacks}
\label{Sec:targeted-attacks}

To better demonstrate the effectiveness of our defense method, we also evaluate our method on targeted attacks, such as AIP \cite{liu2020adversarial} and TAaMR \cite{di2020taamr}.  
Targeted attacks are designed to elevate the positions of specific items within the ranking list generated by a recommender system. In AIP, according to different knowledge the attacker can access, they mainly implement two kinds of attack methods, Insider Attack (INSA) and Expert Attack (EXPA). INSA assumes that the attacker has insider access to the user embeddings of the trained model, reflecting deep insider knowledge. And EXPA is based on the assumption that the attacker can choose a popular item as the point of manipulation (the "hook") and  it also has access to the visual feature extraction model, indicating a high level of expertise and resource access. TAaMR employs two classification-based adversarial attacks, the Fast Gradient Sign Method (FGSM) and the Projected Gradient Descent (PGD), to assess visually-aware recommender systems. The TAaMR attack operates by misclassifying targeted items into the most popular class of items based on the dataset used to train the recommender system.

For the evaluation of targeted attack, we use the prediction shift change, which can show the rank (positive/negative) change before and after attack, following the evaluation metrics for building trustworthy top-N recommender system \cite{mobasher2007toward}.  The average prediction shift $\Delta_{r_i}$ for item i, and the mean average prediction shift for a set of test items $\Delta_{set}$ are defined as follows:   

\begin{equation} \label{eq:metric_shift}
    \Delta_{r_i}=\sum_{u\in\mathcal{U}}\frac{(r_{u,i}^{\prime}-r_{u,i})}{|\mathcal{U}|} \quad\Delta_{\mathrm{set}}=\sum_{i\in\mathcal{I}_{\mathrm{test}}}\frac{\Delta_{r_i}}{|\mathcal{I}_{\mathrm{test}}|},
\end{equation}
where $r'_{u,i}$ indicates the score of user $u$ towards item $i$ given by post-attack recommendation model and $r_{u,i}$ is the prediction score given by a normal recommendation model. Let $r''_{u,i}$ denote the score generated by our defense model. When $r'_{u,i}$ is replaced with $r''_{u,i}$, $\Delta_{set}$ can measure the change in rank before and after applying the defense.

\begin{table}[t]
\caption{Performance (mean average prediction shifts) of our method against targeted attacks on Amazon Men dataset w.r.t. different values of attack perturbation level $\epsilon$.}
  \label{tab:target_attack}
	\begin{tabular}{lcccc}
	    \hline
		attack/defense & $\epsilon=4$ & $\epsilon=8$ & $\epsilon=16$ &$\epsilon=32$ \\\hline
            AIP-INSA  &1.2434	&2.8743	&4.2543	&7.1267  \\
            AIP-INSA (defense) &0.0231	&0.0832	&0.1217	&0.3221 \\ \hline
            AIP-EXPA &1.1864	&2.8322	&5.2955	&8.0270 \\
            AIP-EXPA (defense) &0.0220	&0.0712	&0.1204	&0.3045 \\ \hline
            TAaMR-FGSM &2.5693	&3.9821	&6.3798	&10.5739 \\
            TAaMR-FGSM (defense) &0.1203	&0.2237	&0.3276	&0.5378 \\ \hline
            TAaMR-PGD &3.1293	&4.2398	&7.3749	&11.672 \\
            TAaMR-PGD (defense) &0.2652	&0.4319	&0.9833	&1.5822 \\
		\hline
	\end{tabular}
 \vspace{-10pt}
\end{table}

Table \ref{tab:target_attack} shows the results using the metric of mean average prediction shifts. Two main targeted attack methods we explore here are AIP \cite{liu2020adversarial} and TAaMR \cite{di2020taamr}. The dataset we use for performance comparison is Amazon Men. We also try different attack perturbation magnitude from $\epsilon = $\{4, 8, 16, 32\}. In all the implementations presented in Table \ref{tab:target_attack}, we employed the VBPR model as the foundational backbone of the recommender system. There are three specific attack methods in AIP. Except for the semantic attack which edits the original image using software like Photoshop, we implemented the other two types of attacks, INSA and EXPA. TAaMR uses FGSM and PGD to generate the adversarial perturbations. We defend the recommendation model using our proposed framework against each attack method. In terms of the definition of prediction shifts, larger positive values indicate more successful attacks, which rank the target items higher. Conversely, smaller positive values mean our method can safeguard the RS to give prediction score that closely approximates the original one. To observe the obvious difference, we do not scale the prediction score into 0-1 range.

\section{Discussion and CONCLUSION}

In this work, we present an \emph{adversarial image denoising and detection} framework, that combines two defense strategies (robust model construction and attack detection), to defense against adversarial attacks to visually-aware  recommender systems.  The proposed framework consists mainly of two components: a denoising network for generating clean images and a detection network for separating adversarial inputs from clean inputs. Two networks play different roles in the defense against different magnitudes of adversarial attack signals.  Extensive experiments validated the performance of the proposed framework in defensing and detecting adversarial attacks.

\emph{Implications for research.} Our research contributes to the emerging literature on AI security by introducing an framework to defense against adversarial attacks to visually-aware  recommender systems. As AI-powered recommendation systems being widely used in a wide range of Internet services, security of recommender systems  has received increasingly more attention in recent years. 
While the focus of securing recommender systems has been on developing robust recommender systems, we show that the robustness be been further enhanced by the detection of adversarial inputs. 

\emph{Practical Implications.}  Our study has important implications. As  images are becoming more widely associated with items, visually aware recommender systems (VARS) have been widely used in several application domains such as e-Commerce (\eg Amazon) and  social media services (\eg Pinterest). However, VARS are vulnerable to item image adversarial attacks. As we show in the empirical study, our defense framework is effective in defensing both untargeted and targeted attacks. Furthermore, our framework can be trained and used as a pre-processing step without affecting the currently running recommender system.
Those platforms can use our defense method to enhance the security of VARS.

\emph{Limitations and Future Research.} There are a few limitations in this work and interesting future research
directions. 
First, in this study, we have only experimentally tested two most well-known attack models, namely FGSM and PGD on two popular datasets (Amazon Man and Fashion). An interesting future work is to investigate defenses against different attacks, and test the performances on different datasets other than Amazon dataset.  
Second, the rich interaction between denoising and the detection network can be further exploited by joint optimization of two modules. On the one hand, denoising can benefit detection because a defense system can strategically manipulate adversarial perturbation so that the objectives of cooperating with both the recommender system and the detection systems can be met. On the other hand, adversarial contrastive learning can also facilitate the task of image denoising, because they share the common interest of distinguishing adversarial attacks (aiming at recommender systems) from innocent perturbations such as JPEG compression. The topic of how to achieve self-supervised training of the joint denoising and detection network is left for our future study. 
Another interesting future work is the study of multi-modality attacks (\eg attacks that manipulate both user-item interaction data and the images associated with items) to recommendation systems and their defenses.

\begin{acks} 
This work is partially supported by the NSF under grants CMMI-2146076, CNS-2125958, CCSS-2348046, and CNS-2125977. The authors would like to thank the anonymous reviewers for their valuable comments and suggestions to improve the quality of the article.
\end{acks}

\bibliographystyle{ACM-Reference-Format}
\bibliography{refs}

\end{document}